\documentclass[10pt]{article}
\usepackage{microtype}
\usepackage{verbatim}
\usepackage[left=2.5cm,
right=2.5cm,
top=2.3cm,
bottom=2.3cm,
headheight=20pt,
headsep=10pt,
footskip=25pt,
letterpaper]{geometry}
\usepackage[utf8]{inputenc}
\usepackage[T1]{fontenc}
\usepackage[english]{babel}
\usepackage{amsmath,amsfonts,amssymb,amsthm,thmtools}
\usepackage{graphicx}
\usepackage{float}
\usepackage[ruled]{algorithm}
\usepackage[hyperfootnotes=false]{hyperref}
\usepackage{fontawesome5}
\usepackage{fancyhdr}
\usepackage[normalem]{ulem}
\usepackage{graphicx}
\usepackage{stfloats}
\usepackage{wrapfig}
\usepackage{epstopdf}
\usepackage{cleveref}
\usepackage{subfloat}
\usepackage{subcaption}
\usepackage{xspace}
\usepackage{enumitem}
\usepackage{listings}
\usepackage{titlesec}
\usepackage{etoolbox}
\usepackage{setspace}
\usepackage{changepage}
\usepackage{etoolbox}
\usepackage{multirow}
\usepackage{booktabs}
\usepackage{tabularx}
\usepackage{colortbl}
\usepackage{wrapfig}
\usepackage{svg}
\usepackage[percent]{overpic}
\usepackage{tikz}
\usetikzlibrary{arrows.meta,calc,positioning}
\usepackage[round]{natbib}
\usepackage[colorinlistoftodos, shadow,color=blue!30!white
]{todonotes}
\usepackage{placeins}
\usepackage{xpatch}
\usepackage{siunitx}
\usepackage[htt]{hyphenat}
\usepackage[most]{tcolorbox}

\fancypagestyle{first}{\fancyfoot[R]{\small\thepage}}

\setlist[itemize]{leftmargin=1em,itemsep=0ex,topsep=0ex}
\titlespacing*{\paragraph}{0pt}{0ex plus .1ex}{1ex}
\titlespacing*{\section}{0ex}{2.3ex plus .3ex minus .0ex}{.6ex plus .3ex minus .2ex}
\titlespacing*{\subsection}{0ex}{1.5ex plus .3ex minus .5ex}{.4ex plus .2ex minus .1ex}
\titlespacing*{\subsubsection}{0ex}{1.2ex plus .3ex minus .3ex}{.3ex plus .2ex minus .2ex}

\xapptocmd\normalsize{%
\abovedisplayskip=.8em plus .2em minus .2em
\belowdisplayskip=.6em plus .1em minus .1em
\abovedisplayshortskip=.8em plus .2em minus .2em
\belowdisplayshortskip=.6em plus .1em minus .1em
}{}{}

\setcitestyle{authoryear,round,aysep={,},yysep={,}}
\renewcommand{\cite}[1]{\citep{#1}}

\definecolor{medalgold}{HTML}{FDEBB2}
\definecolor{medalsilver}{HTML}{E4E7EB}
\definecolor{medalbronze}{HTML}{F0D8C0}

\definecolor{mydarkblue}{rgb}{0.0,0.15,0.7}
\definecolor{figureink}{HTML}{212121}
\definecolor{figuremuted}{HTML}{6E6E6E}
\definecolor{figureblue}{HTML}{3A86FF}
\definecolor{figurenavy}{HTML}{264653}
\definecolor{figureorange}{HTML}{FB5607}
\definecolor{figureviolet}{HTML}{5E56E8}
\definecolor{figuregreen}{HTML}{2A9D8F}
\definecolor{figurered}{HTML}{E11D48}
\definecolor{figureamber}{HTML}{F0A202}
\definecolor{figurebluefill}{HTML}{EFF5FF}
\definecolor{figurenavyfill}{HTML}{F2F4F5}
\definecolor{figureorangefill}{HTML}{FFF7F3}
\definecolor{figurevioletfill}{HTML}{F6F6FD}
\definecolor{figuregreenfill}{HTML}{F2F9F8}
\definecolor{figuregrayfill}{HTML}{FAFAFA}
\tikzset{
  case card/.style={rounded corners=3pt, line width=0.7pt, fill=white},
  case band/.style={rounded corners=3pt, line width=0.7pt, draw=figuremuted, fill=figuregrayfill},
  case flow/.style={-{Latex[length=3.2pt,width=4.2pt]}, line width=1.05pt, draw=figureink},
  case feedback/.style={-{Latex[length=2.8pt,width=3.8pt]}, line width=0.75pt, draw=figuremuted},
  case secondary/.style={-{Latex[length=2.6pt,width=3.4pt]}, densely dashed, line width=0.65pt, draw=figureink},
  case heading/.style={font=\sffamily\bfseries\fontsize{8.5}{9.5}\selectfont, align=center},
  case body/.style={font=\sffamily\fontsize{6.5}{7.5}\selectfont, text=figureink},
  case small/.style={font=\sffamily\fontsize{5.8}{6.7}\selectfont, text=figuremuted},
}
\definecolor{abstractfill}{HTML}{F2F3F8}
\definecolor{abstractframe}{HTML}{DADEE9}
\newtcolorbox{abstractpanel}{%
  enhanced, breakable,
  colback=abstractfill, colframe=abstractframe,
  boxrule=0.7pt, arc=11pt,
  left=18pt, right=18pt, top=11pt, bottom=11pt, boxsep=0pt,
}

\hypersetup{%
colorlinks=true,
linkcolor=mydarkblue,
citecolor=mydarkblue,
filecolor=mydarkblue,
urlcolor=mydarkblue}

\makeatletter

  \renewcommand{\maketitle}{%
    \begingroup
      \setlength{\parindent}{0pt}%
      \raggedright
      \vspace*{-1.1cm}%
      \includegraphics[height=.8cm]{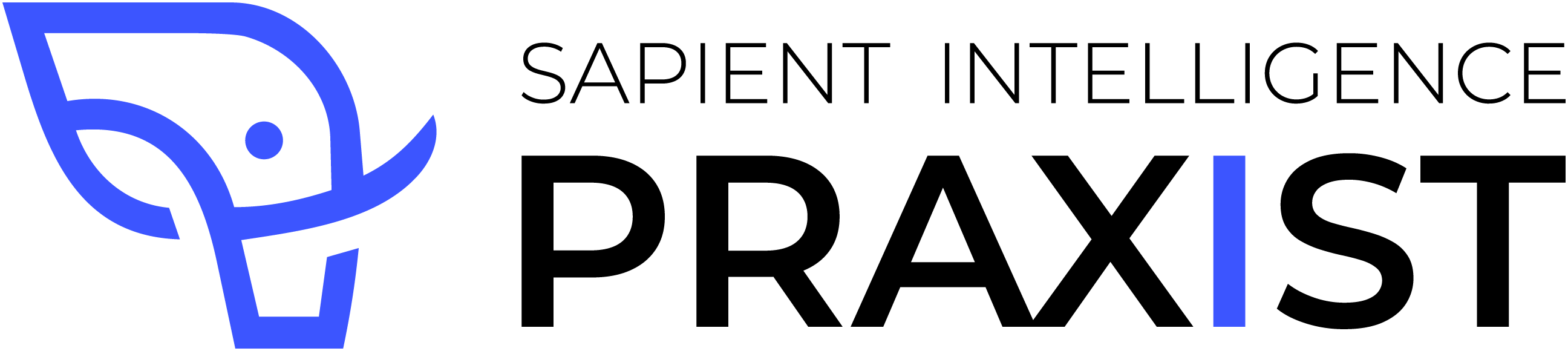}\par
      \vskip .7em
      {\LARGE\bfseries\@title\par}%
      \vskip 0.75em
      {\@author\par}%
      \vskip 0.28in minus 0.1in
    \endgroup
  }
\makeatother

\usepackage{algpseudocode}
\usepackage{pifont}
\usepackage{longtable}
\usepackage{placeins}
\newcommand{\gpupct}{52\%}
\newcommand{\dspct}{43\%}
\newcommand{\subpct}{4\%}
\newcommand{\totalcostv}{\$7,703}

\newcommand{\tokensv}{88}

\newcommand{\gpuhv}{2,020}
\newcommand{\dsusdv}{\$3,338}
\newcommand{\dsprov}{\$3,225}
\newcommand{\dsflashv}{\$114}
\newcommand{\gpucostv}{\$4,040}

\definecolor{cpraxist}{HTML}{2152C4}
\definecolor{coperator}{HTML}{2F86CC}
\definecolor{chuman}{HTML}{0B2A5B}
\definecolor{cgreen}{HTML}{0B919A}
\definecolor{cgray}{HTML}{5B6B82}
\definecolor{rowhl}{HTML}{DCE6FA}
\newcommand{\px}{\textsc{Praxist}\xspace}
\newcommand{\pxd}{\mbox{\textsc{Praxist-RA}}\xspace}
\newcommand{\vc}{VenusCell\xspace}
\newcommand{\fig}[1]{Figure~\ref{#1}}
\newcommand{\tab}[1]{Table~\ref{#1}}
\newcommand{\secref}[1]{Section~\ref{#1}}

\usetikzlibrary{fit,backgrounds,shapes.misc}

\title{Human-Steered Recursive Improvement in a Month-Long Virtual Cell Research Campaign}

\author{
    Zhiyu Wang$^{1}$, Jinling Yuan$^{2}$, Bingxin Zhou$^{2}$, Jin Li$^{1,\dagger}$
  \\[1.2ex]
  $^{1}$Sapient Intelligence \qquad $^{2}$Shanghai Jiao Tong University
  \\[1.0ex]
  \href{https://github.com/sapientinc/praxist}{\faGithub\;\texttt{github.com/sapientinc/praxist}}
}

\begin{document}
\pagestyle{fancy}
\maketitle
\thispagestyle{first}
\vspace{-0.6em}
\begingroup\renewcommand\thefootnote{\fnsymbol{footnote}}\footnotetext[2]{Corresponding author.}\endgroup

\begin{abstractpanel}
{\noindent\textbf{Abstract.}\hspace{0.5em}Autoresearch systems, which run experiments and improve their own results, have advanced mostly where every attempt can be scored quickly and cheaply, in runs of days and with no people after launch. Real research differs on all three counts, since it runs for months, its true evaluation is expensive and rare, and people take part throughout. We therefore ask whether autoresearch can keep improving over a month and what sustains it, whether it can work when true evaluation is scarce, and what human involvement it needs. To study these questions, we propose \emph{human-steered recursive improvement} and implement it in \pxd, a research assistant that couples a general human-interaction harness with the \px autoresearch harness. A human team steers and edits both harnesses, while the agent harness manages an evolving local proxy that stands in for the scarce official evaluation. We ran this system for a month on the Virtual Cell Challenge 2026{\renewcommand\thefootnote{\fnsymbol{footnote}}\footnotemark[1]}, which has a hidden validation grader that allows two submissions a day and a held-out test grader that scores each team's final entry once, when the challenge closes. The resulting model, \vc, raised the official validation score from 0.1041 to 0.2342, and its final entry ranked in the top 8.4\% (123/1{,}458) on the validation leaderboard and the top 12.6\% (102/807) on the test leaderboard, for an estimated \totalcostv. Along the way, we showed that a local proxy for the rationed official grader can be learned, its correlation with the official score rising from 0.24 to about 0.8, and found that improving this proxy and the harness sustained the rise in score. We further analysed human involvement systematically and found that interpreting the official results helped more than proposing methods, and that a few short messages a day were enough to steer the campaign.
\par}
\end{abstractpanel}\vspace{-0.3em}
\begingroup\renewcommand\thefootnote{\fnsymbol{footnote}}\footnotetext[1]{\url{https://virtualcellchallenge.org}}\endgroup

\section{Introduction}
\label{sec:intro}

Research itself is increasingly being automated. Autoresearch systems propose ideas, run experiments and improve an artifact against an automatic score, from machine-learning engineering \citep{aide} and model training \citep{autoresearch} to algorithm discovery \citep{alphaevolve} and end-to-end scientific papers \citep{ai-scientist,ai-scientist-v2,kosmos}. Steered by scientists, such systems have produced findings in astronomy, economics and genomics \citep{discoveryagent}. Self-improving systems go a step further and change the agent that does the work, by evolving its context \citep{ace}, optimising its harness \citep{meta-harness} or designing new agents \citep{adas}. The long-standing goal behind this line of work is recursive self-improvement (RSI), in which a system improves the very means by which it improves, so that gains can compound \citep{stop,godel-agent,dgm,hyperagents}. Recent surveys trace how quickly this field has grown \citep{self-evolving-survey,sico-survey}.

\begin{figure}[tb]
\centering
\resizebox{\linewidth}{!}{\input{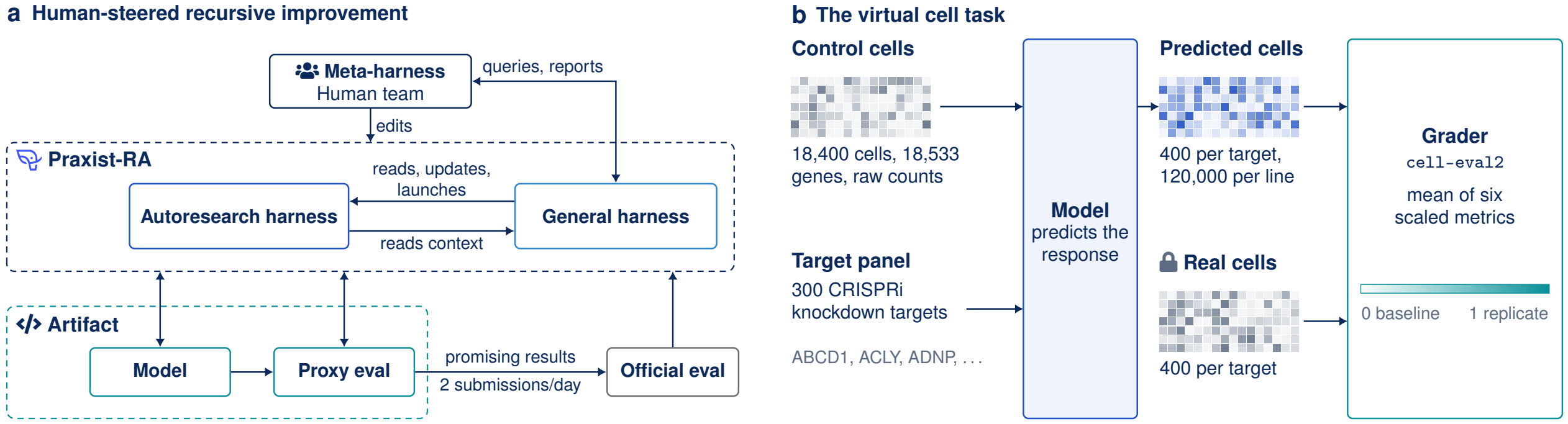}}
\caption{\textbf{\pxd and the Virtual Cell Challenge.} \textbf{a,} \pxd couples a general harness for human interaction with the \px autoresearch harness, and a human team steers and edits both. Together they develop the artifact, a model and its evolving local proxy, and promising results go to the rationed official grader. \textbf{b,} The model turns control cells of an unseen line and a panel of knockdown targets into predicted cells, which the grader compares with hidden real cells. The proxy approximates the grader.}
\label{fig:overview}
\end{figure}

So far, however, this progress rests on three conditions that real research rarely meets. First, every attempt can be scored quickly and cheaply, so the score serves as both objective and guide, whereas the evaluation that matters in research is often expensive and rare. Second, runs last days or at most a few weeks \citep{aide2,dgm,deepscientist,fars}, whereas research projects run for months. Third, people step aside after launch. Most systems take no human input, and the few that do mostly ask a person to rate or choose outputs \citep{sico-survey}, whereas in real research people stay involved throughout, setting directions, interpreting results and deciding where to spend effort. Whether autoresearch can work without these conditions is therefore largely open, and we ask three research questions. RQ1: Can autoresearch keep improving over a month, and what sustains the improvement? RQ2: Can it work when true evaluation is rare and expensive? RQ3: What do people contribute when they stay in the loop, and how much of their involvement is needed?

To study these questions, we propose \emph{human-steered recursive improvement}, in which people steer both how an autoresearch system improves its artifact and how the system itself is improved. We implement it in \pxd, a research assistant built on the autoresearch harness \px \citep{praxist2026}. \pxd couples it with a general human-interaction harness so that a human team can steer the running search through conversation and edit both harnesses as the meta-harness (\fig{fig:overview}a). As a case study, we chose the 2026 Virtual Cell Challenge \citep{vcc2026site,roohani2025vcc}, in which teams predict how unseen human cell lines respond to 300 gene knockdowns (\fig{fig:overview}b), for four reasons. First, true evaluation is scarce. The validation grader returns only two results a day, which tests whether a search can keep improving under rationed evaluation, and a blind test phase scores each team's final entry once on new lines, which checks for overfitting to the validation lines. Second, the challenge stays open for months. Third, the problem is open and free of leakage. The challenge supplies neither training nor validation data for the query lines, and predicting responses in an unseen cell context remains hard \citep{ahlmann2025,systema,pie2026}. Since the responses of the query lines were never published, the backbone LLMs cannot have learned the answers, a known weakness of benchmarks built from public tasks \citep{bench-mlebench,swebench-plus,swebench-illusion}. Finally, a public leaderboard of 1{,}458 teams puts results in context.

We first examined whether autoresearch can keep improving over a month, and what sustains the improvement (RQ1). Over 34 days, the official validation score rose from 0.1041 at the first submission, 608th on the validation leaderboard, to 0.2342 for the final entry, 123rd of 1{,}458 teams and the top 8.4\%. On the blind test lines, the same final entry scored 0.2199 and ranked 102nd of 807 teams, the top 12.6\%. New bests kept arriving without a clear ceiling, and most large steps followed an \emph{improvement of the means of improvement}, such as a repair of local scoring, an update of the harness or more frequent official feedback, rather than a new model. The resulting pipeline, \vc, needed only public data, and the campaign cost about \totalcostv, of which \gpucostv\ paid for GPU hours and \$3{,}663 for LLMs.

We then asked whether autoresearch can work when true evaluation is rare and expensive (RQ2). Autoresearch usually holds its evaluator fixed and trusts it \citep{aide,autoresearch}, whereas we let the agent harness manage an evolving local proxy as part of the artifact. Without experience of the task, a good proxy was hard to build, and its first version, built before any experiment, overstated the official score three times. As official results came in, the agents rebuilt the proxy whenever the two disagreed, and its last version simulated the blind test phase. Later versions reached a Pearson correlation with the official score of up to 0.88, against 0.24 for the first.

Finally, we asked what people contribute when they stay in the loop, and how much of their involvement is needed (RQ3). Existing work has rarely studied sustained human steering of autoresearch systematically \citep{sico-survey,discoveryagent}. To answer RQ3, we therefore kept a full record of the campaign and analysed its human steering, classifying every message of the researcher by the kind of input it gave and tracing the main interventions to what followed. We found that research directions and technical ideas from people helped, but interpreting the official results helped more. For example, the researcher noticed that one of the six metrics of the official score had stayed at zero so far, and the request that followed led to the largest gain of the campaign. We also examined which influences lasted. Rules that could be written down held after one statement, whereas judgments about how to spend compute and effort had to be given again and again, even after a tool was built for them. Even so, a few short messages a day were enough to steer the campaign, with long silences between.

\textbf{Our main contributions are:} (i) We present \pxd, a research assistant that couples a general human-interaction harness with the \px autoresearch harness, so that a human team can steer a running search through conversation and edit both harnesses as it runs, while the general harness keeps the research going through every update and interruption. (ii) We show that human-steered recursive improvement can keep a search improving at the scale of a month, and that its largest gains came from improving the means of improvement. On the Virtual Cell Challenge, new bests kept arriving without a clear ceiling, and the final entry reached the top 8.4\% on validation and the top 12.6\% on the test for about \totalcostv, with all 47 official submissions (Appendix~\ref{app:subs}) and their costs reported. (iii) We propose letting the agent harness manage an evolving local proxy when true evaluation is expensive and infrequent, and show that it works, as the proxy came to track the official score with a correlation of about 0.8. (iv) We analyse human steering systematically over the whole campaign, and find that interpreting the official results helped more than proposing methods, and that rules held once given while judgments about where to spend effort had to be repeated.

\section{Setting}
\label{sec:setting}

We describe the campaign in the vocabulary of self-improving systems. On a task, an agent develops an artifact $y=(T,c)$. The output code $c$ is the program that produces the results, and the task harness $T$ holds everything else the task needs, from the instructions and datasets to the scripts for local evaluation and submission. Autoresearch improves only $c$ under a fixed $T$ \citep{autoresearch,aide}. A system improves itself when experience changes its knowledge, its model or its harness $H$ \citep{ace,meta-harness}, and the improvement is recursive when the means of improving are themselves improved \citep{stop,dgm,hyperagents}. In our setting the harness is \pxd, $H=(H^{\mathrm{G}},H^{\mathrm{A}})$ (\fig{fig:overview}a). The autoresearch harness $H^{\mathrm{A}}$ is \px \citep{praxist2026}, which proposes, builds and evaluates candidate code in parallel and keeps a structured record of the evidence. It works in runs, each a launch of several parallel agents, or peers, that start from the deployed entry and a written brief. The general harness $H^{\mathrm{G}}$ shares its context and serves as the interface for people. It can read and update $H^{\mathrm{A}}$, launch and stop its runs, revise $T$, read the literature and build parts of $c$ itself, and it keeps the reports and lessons that carry knowledge from one run to the next. A human team of one researcher and the \px developers acts as the meta-harness $M$. The researcher steers through conversation with $H^{\mathrm{G}}$, which answers while the search keeps running, and the team also edits both harnesses directly. New versions of the harnesses start from people, whereas changes to $T$ can also come from the agents in pursuit of a goal that a person set. This is how the evolving local proxy arose, since local evaluation belongs to $T$ while the official evaluation stays outside the system. Appendix~\ref{sec:defs} gives the full loop as Algorithm~\ref{alg:praxist}. The backbone models stay frozen throughout. \px used DeepSeek models (V4 Pro and V4.1 Flash) and GPT-6.1 through Codex, and the general harness used Claude Opus 5. Before \px started, a warm-up phase of three days set up the task and built a first baseline with the AutoScientists autoresearch harness \citep{autoscientists}, under the same general harness.

The 2026 Virtual Cell Challenge \citep{vcc2026site,roohani2025vcc} asks teams to predict how unseen human cell lines respond to gene knockdowns. For each query line, the organisers release only unperturbed control cells and a panel of 300 CRISPRi targets, and a submission holds 400 predicted cells per target (\fig{fig:overview}b). The validation phase uses three unnamed lines, A, B and C, and the blind final phase uses three further lines, D, E and F. The grader, \texttt{cell-eval2} \citep{celleval2}, averages six metrics, each scaled so that predicting the mean response of the panel scores 0 and a replicate of the real cells scores 1. Three of them matter most in what follows. The first, \texttt{pds}, asks whether each predicted profile is closest to that of its own target. The second, \texttt{mse}, is the error between the average predicted and real profiles of each target, corrected for sampling noise, and it scores 0 for any error above a fixed threshold. The third, \texttt{nmae}, measures the error on the genes that truly change. The other three score the direction and overlap of differentially expressed genes, and Appendix~\ref{app:score} defines all six. Each team has two validation submissions per day. On this scale the median of the leaderboard is 0.071. Ranks are on the validation leaderboard as frozen on 22 October and published with the final results, with 1{,}458 teams, and results are on validation unless marked as test results, ranked on the final test leaderboard of 807 teams, published on the same day.

The campaign ran for 34 days, from the first official submission just before midnight on 18 September 2026, which we call Day 0, to the freeze of the leaderboard on 22 October, Day 34. Over the campaign, the system ran 4{,}550 experiments, each scoring one candidate on the local proxy, and 47 official evaluations on one 8-GPU node, and the run was about 2$\times$ as long as the longest runs of related systems (\tab{tab:runs} in the appendix). We supplied a corpus of 7 public Perturb-seq datasets in 30 contexts, about 14M cells (\tab{tab:datasets} in the appendix), and the agents were free to extend it. Since the challenge supplies no perturbed cells of the query lines, finding data became part of the research, and by about Day 27 the agents had added 6 further datasets in 7 contexts. Before the search, the general harness built the first version of the local proxy by reimplementing the \texttt{cell-eval2} grader \citep{celleval2} on two lines we held out.

\section{A Month of Improvement}
\label{sec:trajectory}

\begin{table}[b]
\centering\footnotesize
\caption{\textbf{Checkpoints of \fig{fig:score}a.} What the system changed to reach each one, with ranks on the leaderboard.}
\label{tab:checkpoints}
\begin{tabularx}{\linewidth}{@{}rrrrX@{}}
\toprule
CP & Day & Score & Rank (top) & What changed\\
\midrule
1 & 1 & 0.1081 & 596 (40.9\%) & Averaged the responses observed in related public cell lines (warm-up baseline).\\
2 & 5 & 0.1323 & 517 (35.5\%) & Weighted each public cell line by its similarity to the target line, and enlarged the predicted effects.\\
3 & 9 & 0.1787 & 291 (20.0\%) & Predicted the size of each response separately from its direction, the first credit for size.\\
4 & 16 & 0.1934 & 235 (16.1\%) & Enlarged the part of each prediction specific to its target, by a factor set from the line's own controls.\\
5 & 18 & 0.2046 & 192 (13.2\%) & Weighted each prediction by the amount of evidence behind it.\\
6 & 22 & 0.2170 & 160 (11.0\%) & Ranked predictions by noise-corrected confidence, and shared response sizes among similar targets.\\
7 & 25 & 0.2224 & 145 (9.9\%) & Searched for and added new public datasets, among them the Pfizer screens \citep{wong2026}.\\
8 & 31 & 0.2344 & 122 (8.4\%) & Retrained every learned part, then the generative model of single cells, with more public cell lines.\\
\bottomrule
\end{tabularx}
\end{table}

\begin{figure}[t]
\centering
\includegraphics[width=\linewidth]{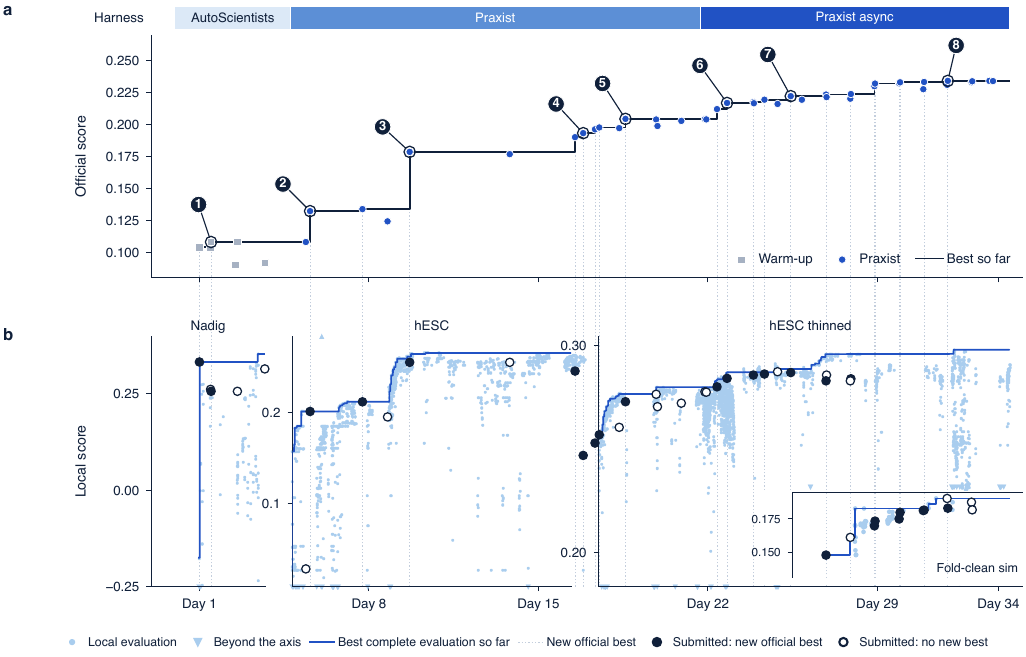}
\caption{\textbf{The campaign on one time axis: harness, official score and local exploration.} \textbf{a,} Official score of all 47 submissions and the cumulative best, with the autoresearch harness above. Numbers mark the notable findings that drove the improvement, CP1--CP8, listed in \tab{tab:checkpoints}. Dotted lines mark each new official best and continue into panel b to the corresponding local submission. \textbf{b,} Logged local evaluations on the version of the local proxy that judged them, each version with its own scale, and, inset, candidates additionally screened on the fold-clean simulation. From Day 28, submitted entries were trained on the line that \textit{hESC thinned} holds out, so we ringed them only in the inset.}
\label{fig:score}
\end{figure}

We first asked whether autoresearch can keep improving over a month, and what sustains the improvement (RQ1). \fig{fig:score} shows the campaign on one time axis. \fig{fig:score}a follows the official score through all 47 submissions, with eight checkpoints, CP1--CP8, at the most important improvements (\tab{tab:checkpoints}) and the autoresearch harness that ran the search above it, and \fig{fig:score}b shows the local evaluations behind them. Together they show that the local proxy that judged the artifact and the autoresearch harness both improved along with the artifact, while the official score rose from 0.1041 to 0.2344 and the rank from 608th (41.7\%) to 122nd (8.4\%), without a clear ceiling. To identify what sustained this improvement, we analyse the research trajectory of the campaign empirically, here and in more detail in Appendix~\ref{sec:route}, since a campaign of this length cannot be repeated as a controlled study of its many causes. We find that most large steps followed an improvement of the means of improvement rather than of the model (\secref{sec:contrib}).

\begin{figure}[t]
\centering
\includegraphics[width=\linewidth]{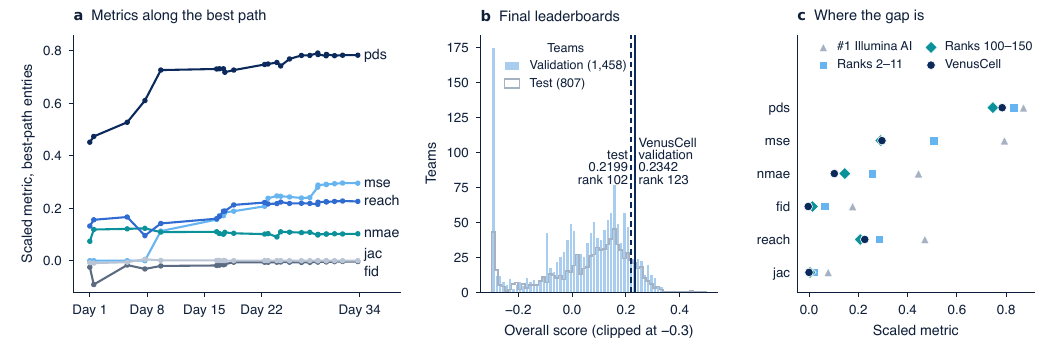}
\caption{\textbf{Score metrics and the final leaderboards.} \textbf{a,} Scaled metrics of the best-path entries. \textbf{b,} All 1{,}458 validation and 807 test scores, clipped at $-0.3$. \textbf{c,} \vc, the top entry and two rank bands on validation.}
\label{fig:members}
\end{figure}

\subsection{The Research Trajectory}
\label{sec:researchtraj}

The campaign opened with a warm-up phase from Day 0 to Day 3, which set up the task and built a baseline with the AutoScientists autoresearch harness \citep{autoscientists}. From the datasets that we supplied, the system built the data pipeline, the submission flow and the first two versions of the local proxy, which Appendix~\ref{app:scorers} defines together with every later version. It then ran 114 experiments across five families of methods, from a generative model of single cells to the transfer of observed responses and learned corrections on top of it. This is more than the longest runs reported by \citet{autoscientists}, which reach about 100 experiments of five minutes each. The first submission scored 0.1041 on Day 0, and the best entry of the warm-up, CP1, averaged the responses observed in related public cell lines and scored 0.1081, 596th and the top 40.9\%. CP1 became the baseline of the campaign, which every later candidate had to beat.

We switched the autoresearch harness to \px \citep{praxist2026} on the evening of Day 4. In its first run, eight parallel agents made 174 evaluations over 13 hours and found that weighting the public cell lines by their similarity to the target line made the predicted response to each target easier to tell apart from those of the other targets, which the \texttt{pds} metric measures, and gave CP2 on Day 5. Like entries of the warm-up before it, this entry gained far more locally than officially, here six times as much, and \pxd traced such gaps to two problems of the proxy. A local gain could include credit that the grader does not pay, and, as the researcher had pointed out, the \texttt{mse} metric had stayed at 0 in every entry (\fig{fig:members}a). Because \texttt{mse} is set to 0 for any error above a fixed threshold, all entries tied on it, and the search could not tell whether a change made the predicted sizes better or worse. On Day 6 the system therefore revised the proxy in two ways. It strengthened the tie between local scores and official results by scaling local scores with the official reference points for 0 and 1 rather than those of its own panel, and it added the raw error behind \texttt{mse} as an objective of its own, so that the search could see a better size of response before the official \texttt{mse} metric moved. \secref{sec:proxies} follows this and every later revision of the proxy and measures how closely each version tracked the grader. With this second signal, the search reached CP3 on Day 9, the largest step of the campaign (+0.0448), which moved the entry from 508th (34.8\%) to 291st (20.0\%). This entry combined three components that the search had developed separately, a model of the size of each response, a model of its direction and a way of generating cells that keeps the predicted average response of each target exact, and Appendix~\ref{sec:method} describes each of them as part of the final pipeline of \vc.

A week without a new best followed, with only one official result, because at that stage each submission still needed the researcher's approval, as the earlier prompts specified, and the agents held candidates back until one looked clearly better. On Day 12 the researcher asked the system to read the literature and public code while the runs continued, one of the human interventions that \secref{sec:human} analyses. Among this code was \texttt{cell-eval2} \citep{celleval2}, the public package that defines the official metrics, and reading it showed the general harness that the predictions understated how much the targets differ from one another on the challenge lines, a common weakness of models that mostly capture the response shared by all targets \citep{systema}. The system therefore enlarged the part of each prediction that is specific to its target, by a factor set for each line from its own control cells and thus without any label, which gave CP4 on Day 16. The first entry built on this finding gained 0.0115 officially yet lost 0.010 on the proxy. Some disagreement was expected, since the proxy held out a stem cell line unlike the challenge lines, but surprisingly \pxd traced this one to a plainer cause, the depth at which the cells were measured, and \secref{sec:proxies} follows how matching that depth gave the version of the proxy that tracked the official grader most closely.

From Day 16 to Day 25, eight further new bests raised the score to 0.2224 and the rank to 145th (9.9\%), and the search widened as it went. The first 15 runs of \px, up to Day 17, introduced 10 new families of related changes, and the next 14 runs, up to Day 28, introduced 37. CP5 on Day 18 came from almost a day without any human message, in which the agents combined four refinements into one entry, among them an idea that had lost officially in the warm-up, pulling each predicted change toward zero when the evidence behind it is weak, which now returned as a weight set by the evidence for each target. After five submissions that each fell short of CP5 by less than 0.006, a major update of \px made research asynchronous on Day 21. On the same day the researcher made a standing rule of using both daily submissions, which the system had mostly done at the researcher's prompting since Day 16. Not every submission carried the highest local score, since many tested whether a single change, such as a stronger or weaker adjustment, held up officially, and the best of many parallel candidates partly reflected the noise of the proxy. New bests then arrived almost daily, with CP6 one day after the update and CP7 on Day 25, and after the update a run was live for 96\% of the time against 76\% before, a gain in continuity that \secref{sec:harness} places among the other improvements of the harness. The data behind CP7 came from the agents themselves, who searched public repositories for new datasets.

In the last phase the goal changed, since the final entry would be scored blind on new cell lines. For that phase the best model would learn from every public line, yet a model trained on every line can no longer be judged on the lines that the proxy holds out. At the team's request, the general harness therefore built a \textit{fold-clean simulation} of the final phase, which \secref{sec:proxies} describes, to screen the leading candidates before submission (\fig{fig:score}b, inset). With it, the system could judge models trained on all lines, which gave a first gain on Day 28 and the best entry of 0.2344 at CP8 on Day 31, 122nd (8.4\%). Four later variants of this entry scored within 0.001 of it, which is comparable to the seed noise of the generative model. The last of them, a 16-model ensemble that we chose for its stability on the blind test lines, scored 0.2342 on the validation lines, and because the leaderboard keeps only a team's latest submission, it lists \vc at 123rd rather than 122nd. With an exact per-cell realisation added after the freeze, which leaves the average prediction for each target unchanged, this ensemble scored 0.2199 on the blind test lines, 102nd of 807 teams (12.6\%). Its drop of 0.014 from validation is typical, since the 141 teams that scored at least 0.2 on validation lost a median of 0.012 on test. Its standing fell from the top 8.4\% to the top 12.6\% mainly because teams with low validation scores rarely entered the test phase (\fig{fig:members}b), as Appendix~\ref{sec:testfield} shows by comparing who entered each leaderboard.

\subsection{Autoresearch Harness Improvement}
\label{sec:harness}

Improving the autoresearch harness itself was part of the campaign, and it came from two sources. The first was the human team, which, besides smaller fixes, released two major updates of \px after its open-source release \citep{praxist2026}. On Day 5, the first replaced fixed generations, in which all peers worked in synchronised rounds of a few hours, with continuous research, in which a run kept going without such boundaries. On Day 21, the second made research asynchronous, so that each peer runs its experiments and reads those of the others without waiting for one another. After this update, helped by changes of the general harness described next, a run was live for 96\% of the remaining 13 days, against 76\% of the 13 days before it. The second source was the general harness, which changed how \px ran the task between these updates, at the researcher's request or on its own initiative. For example, when most GPUs sat idle, it added multi-GPU profiles, a job queue and scoring on the GPU, which made a complete evaluation 3.4$\times$ faster, and from Day 21 it launched each new run without waiting for approval. Appendix~\ref{sec:metaharness} lists every change with its cause.

Updating the harness in the middle of a campaign could not, however, be done by installing a new version alone. Every update ended the running search, and new versions met failures that a run of a few days would not reveal, such as storage that failed under concurrent load. Because of these updates, failures, exhausted allowances and integrity problems, only about 10 of 37 launches ran to their planned end. The general harness made these interruptions recoverable. It diagnosed each failure and patched it locally until a fixed version arrived, kept a report for every run and a file of lessons, and read the evidence that the autoresearch harness recorded. From these it wrote the brief with which each new run started, carrying the deployed entry, the bar that a candidate had to clear and the directions already closed, so a new version of \px resumed the research rather than repeating it. Even the updates written by people therefore reached the running campaign through the general harness, and a campaign that spans many versions of its search needs such a layer above the search to carry the research across them. The backend LLM mattered little, since the search ran mostly on DeepSeek V4 Pro and switching to other backends made no clear difference to the score (Appendix~\ref{sec:efficiency}).

\subsection{Contributors to the Improvement}
\label{sec:contrib}

Read against the trajectory, most large steps followed an \textit{improvement of the means of improvement} rather than of the model alone. Each stall of the first two weeks ended with a repair of the local proxy, which gave CP3 once it could see the size of responses and CP4 once it was brought back to the official depth. In both cases the earlier version of the proxy would not have rewarded the change, since a better size of response left the local score unchanged while \texttt{mse} was stuck at 0, and the first entry built on the finding of CP4 read as a local loss at full depth even as it gained officially. The steady gains of CP5--CP7 followed an update of the autoresearch harness and a doubling of official feedback, and the last checkpoint depended on a new version of the proxy that simulated the blind phase. None of these changes was a new model in itself. Each changed what the search could see or how often it received feedback, and the search then found the model changes that the new view rewarded. In this sense the recursion was practical, since the means of improvement were themselves the main objects of improvement. These changes came from people, often through a short message of the kind that \secref{sec:human} examines, and the system carried them out, building the proxies, reading primary sources and, in the runs of the autoresearch harness, turning each new view into the gains of CP2, CP3, CP5 and CP6. The model changes accumulated on one pipeline, and new architectures, among them learned models of up to 190M parameters and Stack, a single-cell foundation model prompted with cells of other lines \citep{stack2026}, did not displace a pipeline that the search had calibrated for weeks.

\fig{fig:members} shows which parts of the score moved, and the gains came in three of the six metrics, one after another. Before Day 8, every gain came from \texttt{pds}, which measures how well the predicted response to each target can be told apart from those of the other targets, and which rose from 0.47 to 0.73 by CP3 as CP2 reweighted the public cell lines. Then \texttt{mse} rose from 0 to 0.30 once CP3 set the size of each response, and \texttt{reach} rose from 0.14 to 0.23 once CP5 set how far each prediction was trusted. The other three metrics barely moved, and because the grader averages the six metrics equally, they held the score down for the whole campaign. As shown in \fig{fig:members}c, the remaining distance to the top of the leaderboard lies mainly in \texttt{mse} (0.30 against 0.51 for the entries ranked 2--11) and \texttt{nmae} (0.10 against 0.26), whereas on \texttt{pds} \vc was close to them (0.78 against 0.83). More than half of the total gain over CP1 had come by CP3 on Day 9, every later step was smaller, and \secref{sec:cost} relates this slowing to spending.

\section{Evaluation under Scarcity}
\label{sec:proxies}

We then asked whether autoresearch can work when true evaluation is rare and expensive (RQ2). We found that under such scarcity the system could build a local proxy of the official grader and improve it as its experiments accumulated. In this campaign, every version of the proxy held out a set of public lines and targets from training, predicted them as new, and scored the predictions with the six metrics of the official grader. The proxy had to carry nearly all the evaluation, since the search ran about 45 complete local evaluations a day against at most two official results. A search optimises whatever it is scored on, so a proxy that misrepresents the official grader steers the whole search in the wrong direction, and each new version therefore answered a problem that the version before could not handle. Appendix~\ref{app:scorers} defines each version in full and records why each one changed.

\begin{figure}[t]
\centering
\includegraphics[width=\linewidth]{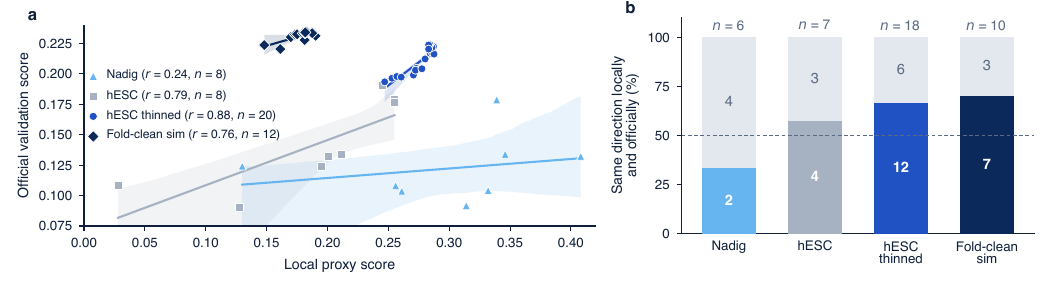}
\caption{\textbf{Local proxies against official results.} \textbf{a,} Local proxy score and official score, with the Pearson $r$ of each version in the legend and 95\% bootstrap intervals as shaded bands. Repeated submissions and submissions without a local score are hidden. \textbf{b,} How often a change from one entry to another moved the local and the official score in the same direction, with counts in the bars.}
\label{fig:proxy}
\end{figure}

The first version held out the HepG2 and Jurkat screens of \citet{nadig2025}, which the team's starting instructions named for validation, and we call it \textit{Nadig}. These lines lie close to the training sources, and all their targets have strong effects, so by Day 1 the general harness had found that this version overstated the official score three times and barely tracked the official grader, as \fig{fig:proxy}a shows ($r=0.24$). On the same day, the researcher asked the system to choose models on a validation of its own and submit only those that looked good enough, and the general harness held out the hESC line of last year's Virtual Cell Challenge \citep{roohani2025vcc}, whose control cells were closest to those of the challenge lines and which, like them, lay far from every training source. This second version, \textit{hESC}, tracked the official grader far better ($r=0.79$) and became the score \px optimised, while \textit{Nadig} became a regression check.

The third version came from a disagreement that the general harness traced to its source. On Day 16, it noticed that \textit{hESC} was reading official gains as local losses, since the first entry built on the finding of CP4 gained 0.0115 officially yet lost 0.010 locally. Investigating the gap, it found that the hESC cells had been counted far more deeply than the challenge cells. A cell is measured by counting its RNA molecules, each tagged with a unique molecular identifier (UMI) \citep{kivioja2012umi}, and the hESC cells carried about 54k UMI each against a median of 20k in the challenge lines, so they were much less noisy. Because the official grader adjusts its scores for noise, the same prediction scored differently at the two depths. The harness therefore randomly discarded counts until the hESC cells matched the challenge depth. This version, \textit{hESC thinned}, ranked four earlier official entries in their official order, replaced \textit{hESC} as the score of \px on Day 17, and tracked the official grader most closely of all versions ($r=0.88$).

In the last few days before the close of the validation leaderboard, the team's goal turned to the blind test phase, in which the final entry would meet new lines. The best model for that phase would learn from every public line, hESC included, so no held-out line remained to judge it. At the team's request, the general harness therefore built a \textit{fold-clean simulation} of the test phase, a leave-one-group-out cross-validation over nine public cancer, normal and stem cell lines, which Appendix~\ref{app:scorers} lists. Each line became a new line in turn, removed with its relatives from the training of every learned part, and three lines were never used to choose a candidate, so that they stood in for the blind test lines. The simulation did not replace \textit{hESC thinned} but worked alongside it. The search kept optimising \textit{hESC thinned}, and the simulation screened its leading candidates before submission, a check that tracked the official score closely ($r=0.76$).

Taken together, we show that the local proxy can be improved recursively as official results accumulate, and that a better proxy in turn made the search better. The correlation between local and official scores rose from 0.24 on \textit{Nadig} to 0.79 on \textit{hESC} from Day 1 and 0.88 on \textit{hESC thinned} 16 days later, so the proxy came to track the trend of the official grader closely and could steer the search toward better models (\fig{fig:proxy}a). Although single comparisons stayed noisy, they improved as well, since the share of changes that the proxy read in the right direction rose from 2 of 6 on \textit{Nadig} to 7 of 10 on the \textit{fold-clean simulation} (\fig{fig:proxy}b).

\section{Contribution of Human Involvement}
\label{sec:human}
\suppressfloats[t]

\begin{figure}[t]
\centering
\includegraphics[width=\linewidth]{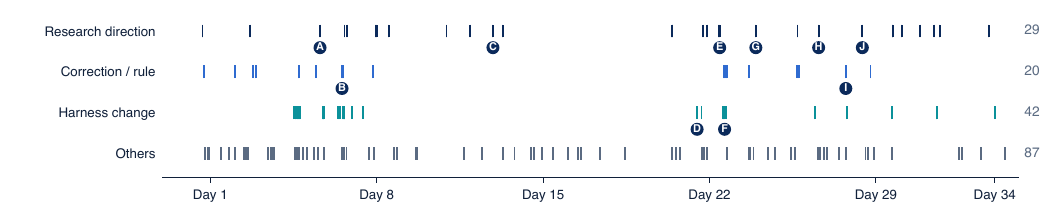}
\caption{\textbf{Steering messages over the campaign.} One tick per message, by class, with counts at the right. Others joins status checks, go-aheads and submission decisions, and letters mark the interventions of \tab{tab:interventions}. Harness messages cluster around the start of \px and the update of Day 21.}
\label{fig:human}
\end{figure}

Finally, we asked what people contribute when they stay in the loop, and how much of their involvement is needed (RQ3). We extracted every message the researcher sent to the general harness up to the freeze of the leaderboard and coded each one by hand, from a status check to steering in conversation \citep{agent-laboratory,tellonce}, with the codebook in Appendix~\ref{app:coding}. Of 302 typed messages, 178 steered the campaign, with a median of 10 words, while the rest concerned administration or reporting (\fig{fig:human}). The steering messages comprise 34 status checks, 21 go-aheads, 32 submission decisions, 20 corrections, 29 research directions and 42 harness changes. Harness changes were direct prescriptions for the system itself, so people also recursively steered how the system improved.

\begin{table}[t]
\centering\footnotesize
\caption{\textbf{Main interventions and what followed.} Messages are paraphrased and grouped by the kind of input, with letters in order of time. Official changes cover later steps, and -- marks an intervention without one.}
\label{tab:interventions}
\begin{tabularx}{\linewidth}{@{}lrp{5.6cm}Xr@{}}
\toprule
& Day & Message & What followed & Official\\
\midrule
\multicolumn{5}{@{}l}{\textit{Reading the official results}}\\
A & 5 & The \texttt{mse} metric is always 0, so lower the error & Raw error as an objective, then the first mse above 0 & +0.0448\\
C & 12 & No gain in two days, so research more widely & Literature track, then a close reading of the grader's source & +0.0191\\
\addlinespace[2pt]
\multicolumn{5}{@{}l}{\textit{Pressure at plateaus}}\\
E & 22 & Look for genuinely novel ideas & Most new methods failed, one size-allocation idea fed run 28 & +0.0048\\
G & 23 & Explain why different methods score alike & New data, where Pfizer screens helped and L1000 failed & +0.0019\\
\addlinespace[2pt]
\multicolumn{5}{@{}l}{\textit{Technical suggestions}}\\
B & 6 & Use GPU methods, such as a generative model & LSA and FM, the direction model of every later entry & --\\
J & 28 & Try Arc's PIE method & Passed all 4 local lines, then lost officially & $-$0.0054\\
\addlinespace[2pt]
\multicolumn{5}{@{}l}{\textit{Rules and resources}}\\
D & 21 & Use both daily submission slots & Official results rose from 1.0 to 2.0 per day & --\\
F & 22 & Switch the backend to GPT-6.1 & Run 28 best, later GPT-6.1 runs set no best & --\\
H & 26 & New datasets may be added to the corpus & Six datasets added, about 60\% of the Day 28 gain & +0.0082\\
I & 27 & GPUs are idle again, so supervise their use & GPU job queue, screens of 100+ cells became routine & --\\
\bottomrule
\end{tabularx}
\end{table}

\tab{tab:interventions} lists the 10 interventions that mattered most, grouped by the kind of input, and three patterns stand out. First, research directions and technical ideas helped (B), but reading the official results helped more (A and C). One might expect a person's main contribution to be new methods \citep{agent-laboratory,discoveryagent}. Instead, in this case study, the most valuable input, intervention A, pointed at one number that the search had never moved. The \texttt{mse} metric was 0 in 108 of 108 complete rows of run 5, and because a metric stuck at 0 cannot show whether a change helps, the search could not see it as a problem. The scientists who steered the system of \citet{discoveryagent} contributed in a similar way, since their interventions mostly reframed objectives, standardised evaluation and interpreted ambiguous findings. Second, generic pressure at plateaus, such as ``be bold'' or ``try something new'' (E and G), was often followed by progress, which in both cases came from a change in what was searched over, such as new data, whereas specific technical suggestions, such as the PIE method \citep{pie2026} (J), did not hold up officially. Third, the researcher acted as a policy maker, setting the rule of Day 21 that put every official slot to use (D) and deciding on resources, from the backend (F) to new datasets (H) and idle GPUs (I).

To see whether the need for steering would shrink as the campaign went on, we next asked which kinds of input stayed in the system once given, grouping the steering messages by theme and checking each group by hand. We found that operational rules held after a single statement. Rules such as submitting nothing unrequested, keeping run state on network storage and never resubmitting an entry to restore a rank could be written down and checked, and none had to be repeated. In contrast, judgments about how to spend compute and effort had to be given again and again. The researcher asked for idle GPUs to be used 14 times on 7 days, four of them after a GPU supervisor and job queue had been built for exactly this purpose, and the request to keep exploring after a local win came about 13 times. The need for steering therefore did not shrink. Although messages per day fell after the first ten days, steering per +0.01 of official score rose from about 11.5 messages early in the campaign to 19.1 at the end, for two reasons that we can measure. Later gains were smaller, with a median of +0.0013 per new best, and judgments such as these had to be given again. In this case study, the system took in rules that it could check, while judgments about where to spend compute and effort still had to come from people \citep{tellonce,co-improvement}. These recurring judgments mark what an automated meta-harness would have to learn before people could step back.

Little involvement was needed while the search ran. The median gap between typed messages was about half an hour, yet most live run time passed in long silences. Message-free stretches longer than 6 hours held 74\% of live run time, and stretches longer than 12 hours held 49\%. The longest stretch without a human message lasted about 21 hours, from early on Day 17 to early on Day 18, and it shows what the system did on its own. In that time, eight peers logged 259 evaluations and stacked four refinements into one composite, which set a new best of 0.2046 the next day as CP5. The system could therefore work productively for about a day without intervention, and the campaign is best read as a month of daily steering over runs that were left to themselves overnight.

Such short messages could steer the campaign because the general harness turned them into work. For example, it worked out from the grader's code how \texttt{mse} corrects for sampling noise, thinned the proxy to the depth of the challenge cells and caught each integrity failure, so that human intent became concrete work for the whole harness. As a general agent, the general harness could also take on requests that no purpose-built interface would anticipate. These results answer RQ3. People contributed judgment above the search, chiefly the interpretation of official results and the decision of when to change course, and a few short messages a day were enough. Since the team came to the campaign with little experience in virtual cell modelling, this suggests that a newcomer to a field can steer the system through judgment, and an expert can later refine it with domain knowledge.

\section{Cost}
\label{sec:cost}

\begin{figure}[t]
\centering
\includegraphics[width=\linewidth]{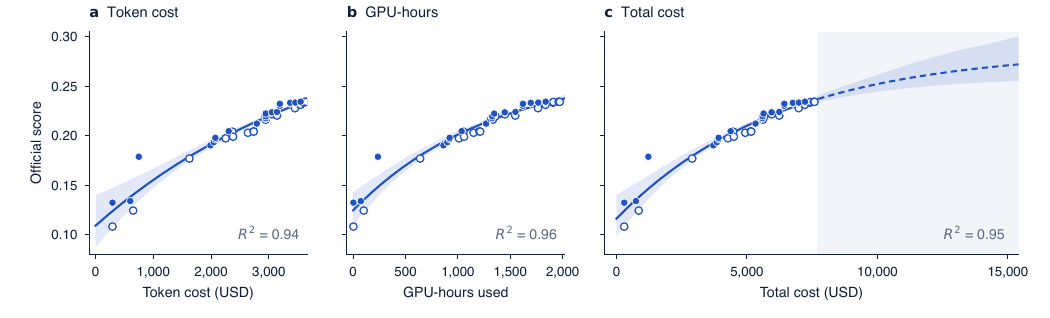}
\caption{\textbf{Score against spending.} \textbf{a,} Token cost: DeepSeek at list price plus one monthly fee for each subscription, charged at its start. \textbf{b,} GPU-hours actually used. \textbf{c,} Total, with GPUs at \$2 per GPU-hour. Each marker is one of the 41 official submissions made with \px\ at its score, filled when it set a new best. The line is a saturating fit to all 41, which rises ever more slowly toward a plateau, shaded with its 95\% bootstrap interval. In c, the grey area doubles the spending past the freeze of the leaderboard, and the dashed line extends the fit into it.}
\label{fig:cost}
\end{figure}

We estimate that the whole campaign cost \totalcostv\ up to the freeze of the leaderboard (Appendix~\ref{app:cost} gives the breakdown). As shown in \fig{fig:cost}, returns fell as the score rose. About \$1{,}220 reached 0.1787, 56\% of the gain, at \$170 per +0.01, and the rest \$1{,}080 per +0.01, six times as much, matching the slowing of the official score in the last weeks. Yet we see no clear ceiling when the leaderboard froze. A saturating fit to the total cost\footnote{We fit $\mathrm{score}=p-g\,e^{-x/k}$ to the 41 submissions, where $x$ is the cumulative cost at submission and $p$, $g$ and $k$ are fitted.} projects a score of about 0.27 at twice the spending, which would rank 66th on the leaderboard. Behind this total are about \tokensv\ billion tokens and \gpuhv\ NVIDIA A100 80GB GPU-hours of use, a modest budget for a month of research on a problem of this difficulty. At \$2 per GPU-hour, close to the median on-demand price of this GPU in 2026, these hours come to \gpucostv, the DeepSeek tokens of the autoresearch harness come to \dsusdv\ at DeepSeek's list price, and we add one monthly fee for the Claude Ultra (\$200) and ChatGPT Team Premium (\$125) subscriptions. The general harness runs on Claude Opus 5 for its widely recognised strength in this role, and a single Claude monthly subscription covered every request to it over the campaign. GPU time accounts for \gpupct\ of the total, DeepSeek tokens for \dspct\ and the two subscriptions for \subpct.

\section{Conclusion}
\label{sec:conclusion}

We studied whether autoresearch can keep improving over a month and what sustains the improvement, whether it can work when true evaluation is scarce, and what people contribute when they stay in the loop, through a campaign of human-steered recursive improvement on the Virtual Cell Challenge. \pxd raised the official validation score from 0.1041 to 0.2342 and reached the top 8.4\% of 1{,}458 teams on validation and the top 12.6\% of 807 teams on the blind test lines. The largest gains followed improvements to the means of improvement rather than new models, an evolving local proxy made a rationed grader workable, and a few short human messages a day supplied the judgment that the search lacked. These results come from a single campaign, selected over 47 submissions of the same validation panel, and on the blind test lines the final entry kept 94\% of its validation score, a drop similar to that of other strong entries. The campaign suggests a practical path toward recursive self-improvement, in which people first carry the meta-harness, the system records what they do, and each recurring judgment then moves inside the loop. We believe that long-term research calls for an autoresearch harness coupled with a general harness, which is the design of \pxd and the direction in which our team continues to develop it.

\clearpage
\bibliographystyle{plainnat}
\bibliography{main}

\clearpage
\appendix
\section{Formal Setting and Loop}
\label{sec:defs}

This appendix gives the formal setting behind \secref{sec:setting}, the loop that the campaign followed and the comparison of its length with other automated research runs. We describe the campaign in the vocabulary of the literature on self-improving systems. An agent is a model with parameters $\theta$ run by a harness $H$, meaning its prompts, tools, context rules and control flow. On a task, the agent develops an artifact $y=(T,c)$, which holds everything specific to that task. The output code $c$ is the program that produces the results, and the task harness $T$ is the rest of what the task needs, namely its instructions, the datasets and the scripts for local evaluation and for submitting results. Each run leaves a trace $\tau$, an external evaluation returns feedback $f$, and persistent knowledge $K$ can carry over to later tasks. Finally, a meta-harness $M$ is whatever rewrites $H$. In these terms, autoresearch improves only $c$ on one problem under a fixed $T$, and the agent ends the run as it started \citep{autoresearch,aide}. A system improves itself when experience changes $K$, $\theta$ or $H$ so that it does better or more cheaply on later tasks of the same kind \citep{ace,meta-harness}. Improvement is \emph{recursive} when the means of improving are themselves improved, so that gains can compound, and recursive self-improvement (RSI) in the full sense asks a system to do this to itself, with $M$ inside the system and among the things that improve \citep{stop,dgm,hyperagents}. One further distinction matters for our setting. The official evaluation lies outside the system, whereas local evaluation is part of $T$, so the system can revise it as the work goes on, whereas it can only spend the official evaluation.

\fig{fig:overview}a shows the setting in three layers. The middle layer is the harness, \pxd, $H=(H^{\mathrm{G}},H^{\mathrm{A}})$, which joins two parts that share one context. The autoresearch harness $H^{\mathrm{A}}$ is \px, a research loop that started from its open-source release \citep{praxist2026} and that proposes, builds and evaluates candidate output code in parallel while keeping a structured record of the evidence. Over it stands the general human-interaction harness $H^{\mathrm{G}}$, the general harness for short, which can read $H^{\mathrm{A}}$ at any moment, from the evidence and logs of a live run to its code and configuration. The general harness can also update $H^{\mathrm{A}}$, launch, stop and resume its runs, revise $T$, read the literature, package submissions, and build and test parts of $c$ on the same GPUs. We report every entry made with \px as the joint output of the two harnesses of \pxd. The top layer is the meta-harness $M$, a human team of one researcher and the \px developers. The bottom layer is the evaluation, where the local scorer in $T$ sees every candidate and promising results go to the official grader, which allows two submissions per day. Both scores return to the shared context. Both harnesses keep knowledge of their own, \px in the evidence record of its runs and $H^{\mathrm{G}}$ in the reports and the numbered lessons file that carry knowledge across runs, and $H^{\mathrm{G}}$ can read the evidence record of \px whenever it needs to (Appendix~\ref{sec:knowledge}).

\begin{algorithm}[t]
\caption{Autoresearch with human-steered recursive improvement. The search and the conversation run concurrently, and every change to $H^{\mathrm{G}}$ or $H^{\mathrm{A}}$ is written by people or prepared by $H^{\mathrm{G}}$ in answer to a human follow-up.}
\label{alg:praxist}
\footnotesize
\begin{algorithmic}[1]
\Require backend $\theta$; harness $(H^{\mathrm{G}},H^{\mathrm{A}})$ (general, autoresearch); artifact $y=(T,c)$ (task harness, output code); knowledge $K$; run budget $b$; human team $M$
\State candidates $C\gets\{c\}$
\State $S\gets\textsc{Launch}(\textsc{Search};\ \theta,H^{\mathrm{A}},T,K,b)$ \Comment{runs in the background}
\While{the challenge is open}
  \State wait for a query $q$ from $M$ \Comment{at any time; $S$ keeps running}
  \State $r\gets H^{\mathrm{G}}\bigl(q,\textsc{Read}(S,H^{\mathrm{A}},T,C),K\bigr)$ \Comment{analyse the search, report back}
  \State $K\gets K\cup\{r\}$, \ $u\gets M(r)$ \Comment{follow-up, possibly none}
  \If{$u$ submits a candidate $c_u\in C$}
    \State $f^{\mathrm{off}}\gets\textsc{Eval}^{\mathrm{off}}(T,c_u)$, \ $K\gets K\cup\{f^{\mathrm{off}}\}$ \Comment{at most 2 per UTC day}
  \ElsIf{$u$ asks for output code}
    \State $C\gets C\cup\{H^{\mathrm{G}}(u,C,T)\}$ \Comment{built and tested by $H^{\mathrm{G}}$}
  \ElsIf{$u$ changes $H^{\mathrm{G}}$}
    \State $H^{\mathrm{G}}\gets$ the version written by $M$ \Comment{applies at once; $S$ keeps running}
  \ElsIf{$u$ changes $T$, $H^{\mathrm{A}}$ or the run settings $(\theta,b)$}
    \State $\Delta\gets$ the version written by $M$, else $H^{\mathrm{G}}(u,H^{\mathrm{A}},T)$ \Comment{prepared while $S$ runs}
    \State \textsc{Stop}$(S)$; apply $\Delta$ to $(T,H^{\mathrm{A}},\theta,b)$
    \State $S\gets\textsc{Launch}(\textsc{Search};\ \theta,H^{\mathrm{A}},T,K,b)$ \Comment{resume, or start a new run}
  \EndIf
\EndWhile
\State \Return the candidate $c^\star\in C$ that $M$ enters in the blind test
\Statex
\Function{Search}{$\theta,H^{\mathrm{A}},T,K,b$} \Comment{$H^{\mathrm{A}}$ and $T$ fixed within a run}
  \While{budget $b$ remains and $S$ is not stopped}
    \State $(C',\tau)\gets\textsc{ResearchLoop}(\theta,H^{\mathrm{A}},T,K,C)$ \Comment{propose, build, record}
    \State $f^{\mathrm{loc}}\gets\textsc{Eval}^{\mathrm{loc}}(T,C')$ \Comment{every candidate}
    \State $C\gets C\cup C'$, \ $K\gets K\cup\{\tau,f^{\mathrm{loc}}\}$ \Comment{visible to $H^{\mathrm{G}}$ at once}
  \EndWhile
\EndFunction
\end{algorithmic}
\end{algorithm}

Algorithm~\ref{alg:praxist} gives the loop, which has two concurrent parts. The first is the search, which runs in the background under $H^{\mathrm{A}}$ for hours to days, with $H^{\mathrm{A}}$ and $T$ held fixed. The second is a conversation between the researcher and $H^{\mathrm{G}}$, which accepts input at any time and leaves the search running while it answers. An exchange starts with a human query, such as how a run is going or why a score moved, and $H^{\mathrm{G}}$ reads the state of $H^{\mathrm{A}}$, analyses it and reports back. The researcher may stop there or follow up. A follow-up can spend an official submission, ask $H^{\mathrm{G}}$ to build part of $c$ itself, or change the system, either through $T$, as with a corrected scorer, a new brief or a further dataset, or through $H^{\mathrm{A}}$, as with a repair or a new capability. $H^{\mathrm{G}}$ prepares such a change while the search continues, and the change takes effect when it stops and resumes the run or launches a new one. Besides the conversation, the team edits both harnesses directly. The \px developers write new versions of $H^{\mathrm{A}}$, which $H^{\mathrm{G}}$ installs, and the researcher edits the standing instructions of $H^{\mathrm{G}}$, such as the procedures for setting up and monitoring a run, which apply at once and leave the search running. New versions of both harnesses come from people, with $H^{\mathrm{G}}$ supplying the analysis before them and the work after them, while $H^{\mathrm{G}}$ can also adjust how $H^{\mathrm{A}}$ runs the task, as with a job queue or a patch for a failure, at the researcher's request or on its own initiative. Changes to $T$, in contrast, can also come from the agents in pursuit of a goal that a person set. A follow-up may give only a goal, in which case $H^{\mathrm{G}}$ designs the change itself, and this is how the local proxies arose, as requests for better results led $H^{\mathrm{G}}$ to compare local and official scores, find why they disagreed and rebuild the scorer.

The setting is therefore recursive because the means of improving were themselves improved. Repairs to the local scorers and to $H^{\mathrm{A}}$ gave later searches sounder evidence, and that evidence fed the reports on which the next changes were decided. The local scorers show this most clearly. They belong to $T$ and so to the artifact, yet every later candidate is judged by them, so revising a scorer changes how the system improves as well as what it produces. The setting is human-steered because the output code improves through the system's own runs while the harnesses change only when a person decides. The role that systems such as DGM and Hyperagents give to a language model \citep{dgm,hyperagents} is thus played here by people assisted by a general agent, and what separates the setting from RSI in the full sense is the seat of $M$, which people hold here and which an automated system would have to take over. Making the human interface a general agent rather than a purpose-built tool is the design choice that makes this practical. A purpose-built interface covers only the situations its designers foresaw, whereas a general agent with conversation, file and shell access, log reading and tool use can improvise when something unforeseen happens, such as a storage failure or a question about the grader's source code. Because it shares the context of the research loop and can inspect it without disturbing it, people can ask questions as often as they like, while the autoresearch harness stays focused on sustained, parallel and evidence-driven search over candidate code.

\begin{table}[t]
\centering\footnotesize
\setlength{\tabcolsep}{3pt}
\caption{\textbf{Long runs of automated research systems.} Each row describes the system's longest run. An external scorer is one outside the authors' team. Open data means the task does not supply all the data, so the run must find its own. For systems with human input, the longest run counts calendar days with people in the loop.}
\label{tab:runs}
\begin{tabularx}{\linewidth}{@{}>{\raggedright\arraybackslash}p{5.0cm}>{\raggedright\arraybackslash}X>{\centering\arraybackslash}p{1.45cm}>{\centering\arraybackslash}p{1.1cm}>{\centering\arraybackslash}p{.75cm}>{\centering\arraybackslash}p{.95cm}>{\centering\arraybackslash}p{.85cm}@{}}
\toprule
System & Benchmark & Longest run (days) & External scorer & Open data & Harness updates & Human input\\
\midrule
The AI Scientist \citep{ai-scientist} & Custom: ML code templates & $<$1 &  &  &  & \\
AI Scientist-v2 \citep{ai-scientist-v2} & Custom: workshop peer review & $<$1 & \ding{51} & \ding{51} &  & \\
Kosmos \citep{kosmos} & Custom: open-ended data analysis & $<$1 & \ding{51} &  &  & \\
AutoScientists \citep{autoscientists} & BioML-Bench \citep{bench-biomlbench} & $<$1 &  &  &  & \\
autoresearch \citep{autoresearch,kahn2026karpathy} & Custom: nanochat training & 2 &  &  &  & \\
DeepScientist \citep{deepscientist} & RAID \citep{bench-raid} & 14 &  &  &  & \ding{51}\\
FARS \citep{fars} & Custom: 67 AI/ML topics & 17 & \ding{51} & \ding{51} &  & \\
\textbf{This campaign} & Virtual Cell Challenge \citep{vcc2026site} & \textbf{34} & \ding{51} & \ding{51} & \ding{51} & \ding{51}\\
\bottomrule
\end{tabularx}
\end{table}

\tab{tab:runs} places the campaign among the longest reported runs of automated research systems. Its 34 days are about twice the 17 days of FARS, the longest run that we found, and it is the only run in the table whose harness changed while it ran. The length column counts different things for different systems, since a run with people in the loop is measured in calendar days. In this campaign a person checked in at least once a day, and the longest stretch of live run time without a human message lasted about 21 hours (Appendix~\ref{sec:unattended}), so the table compares the length of a campaign steered with daily attention rather than the length of unattended operation.

\section{The Virtual Cell Challenge}
\label{sec:vcc}

This appendix describes the task of \secref{sec:setting} in more detail, defines the six metrics of the official grader, and explains how we read ranks on the two leaderboards. An AI virtual cell is a model that predicts how the state of a cell changes under an intervention \citep{bunne2024virtualcell}. Perturb-seq made this question measurable by reading out pooled CRISPR screens at single-cell resolution \citep{dixit2016perturbseq,replogle2022,nadig2025,orion2025}. Models for it range from latent-shift and autoencoder approaches \citep{scgen,cpa} and gene-graph models \citep{gears} to optimal-transport and flow-based models \citep{cellot,cellflow} and single-cell foundation models \citep{geneformer,scgpt,state2025}. Yet simple baselines often match them \citep{ahlmann2025}, and common metrics can reward systematic variation, the response shared across perturbations, over target-specific effects \citep{systema}. The hardest setting is a new cell context, in which no perturbed cell of the line has been observed, and the Virtual Cell Challenge \citep{roohani2025vcc} makes this setting an open competition with a hidden test.

\begin{figure}[t]
\centering
\resizebox{\linewidth}{!}{\newcommand{\ovsee}{\smash{\textcolor{cgreen}{\faEye}}\;}%
\newcommand{\ovlock}{\smash{\textcolor{cgray}{\faLock}}\;}%
\newcommand{\ovrow}[5]{%
  \node[rl] at (0,#1) {#2};
  \node[ce] at (4.925,#1) {#3};
  \node[ce] at (10.075,#1) {#4};
  \node[ce] at (15.225,#1) {#5};
}%
\begin{tikzpicture}[x=1cm,y=1cm,
  bx/.style={rounded corners=1.6pt,draw=#1,line width=.5pt,fill=white,align=center,font=\sffamily\fontsize{6.1}{7.3}\selectfont,text=chuman,inner sep=2.5pt},
  grp/.style={rounded corners=2.5pt,dash pattern=on 2pt off 1.6pt,line width=.45pt,color=#1},
  tt/.style={font=\sffamily\bfseries\fontsize{6.6}{7.9}\selectfont,text=chuman,align=center},
  tl/.style={font=\sffamily\bfseries\fontsize{6.6}{7.9}\selectfont,text=chuman,anchor=west,inner xsep=0},
  tx/.style={font=\sffamily\fontsize{6.1}{7.3}\selectfont,text=chuman,align=center},
  xl/.style={font=\sffamily\fontsize{6.1}{7.3}\selectfont,text=chuman,anchor=north west,align=left,inner xsep=0,inner ysep=0},
  sm/.style={font=\sffamily\fontsize{5.6}{6.7}\selectfont,text=cgray,align=center},
  lb/.style={font=\sffamily\fontsize{5.6}{6.7}\selectfont,text=chuman,align=center,inner sep=1.6pt},
  pl/.style={font=\sffamily\bfseries\fontsize{8}{9.6}\selectfont,text=chuman,anchor=west,inner xsep=0},
  ar/.style={-{Latex[length=2.7pt,width=3.1pt]},line width=.5pt,color=#1},
  dar/.style={{Latex[length=2.7pt,width=3.1pt]}-{Latex[length=2.7pt,width=3.1pt]},line width=.5pt,color=#1},
  rl/.style={font=\sffamily\bfseries\fontsize{6.1}{7.3}\selectfont,text=chuman,anchor=west,align=left,inner sep=0,execute at begin node={\vphantom{Ag}},execute at end node={\vphantom{Ag}}},
  ce/.style={font=\sffamily\fontsize{6.1}{7.3}\selectfont,text=chuman,align=center,text width=5cm,inner sep=0,execute at begin node={\vphantom{Ag}},execute at end node={\vphantom{Ag}}},
  hd/.style={font=\sffamily\bfseries\fontsize{6.6}{7.9}\selectfont,text=#1,text height=1.7ex,text depth=.45ex}]
\useasboundingbox (0,0) rectangle (17.78,3.12);
\draw[cpraxist,line width=1pt] (2.4,3.02) -- (7.45,3.02);
\draw[coperator,line width=1pt] (7.55,3.02) -- (12.6,3.02);
\draw[chuman,line width=1pt] (12.7,3.02) -- (17.75,3.02);
\node[hd=cpraxist] at (4.925,2.87) {Train};
\node[hd=coperator] at (10.075,2.87) {Validation};
\node[hd=chuman] at (15.225,2.87) {Test};
\foreach \y in {.56,.88,1.2,1.52,1.84,2.4,2.72}{\draw[cgray!30,line width=.3pt] (0,\y) -- (17.75,\y);}
\ovrow{2.56}{Cell lines}{HEK293T, HCT116, K562, KOLF, hESC and others}{A, B, C (unnamed)}{D, E, F (unnamed)}
\ovrow{2.12}{Supplied by}{Our team:~7 public datasets, 30 contexts.\\Agents added 6 datasets, 7 contexts}{Organisers}{Organisers}
\ovrow{1.68}{Control cells}{\ovsee 0.70M}{\ovsee 18,400 per line}{\ovsee 18,400 per line}
\ovrow{1.36}{Perturbed cells}{\ovsee 13.3M}{\ovlock 120,000 per line, hidden}{\ovlock 120,000 per line, hidden}
\ovrow{1.04}{Targets}{Differ by screen}{300}{300, of which 129 in validation}
\ovrow{.72}{Nearest other line}{Cosine 0.84--0.92}{Cosine 0.80--0.84}{Cosine 0.75--0.79}
\ovrow{.28}{Feedback}{Local proxy on held-out lines,\\4,550 evaluations}{Official score, 2 per day,\\47 entries}{None:~one blind entry}
\end{tikzpicture}%
}
\caption{\textbf{Data partitions.} Training data come from public screens of other lines, while validation and test lines release only their control cells. Cell counts in the training column cover the 30 starting contexts. Cosine similarity compares each line's controls with the nearest other public line, and the training range spans the six main corpus lines.}
\label{fig:partitions}
\end{figure}

For each query line, the organisers release 18{,}400 unperturbed control cells with raw counts over 18{,}533 genes and a panel of 300 CRISPRi knockdown targets, and a submission contains 400 predicted cells per target (\fig{fig:partitions}). No perturbed cell of a query line is ever visible. The validation phase uses three unnamed lines, A, B and C, and the final phase three further lines, D, E and F, whose panels share 129 targets with the validation panel. All six lines are far from public data. Compared by their control cells, the nearest public line has a cosine similarity of 0.80--0.84 for A, B and C and of 0.75--0.79 for D, E and F, against 0.84--0.92 between the main lines of the training corpus, so matching a query line to a similar known line does not reach it.

\subsection{Scoring}
\label{app:score}

The grader, \texttt{cell-eval2} \citep{celleval2}, averages six metrics, each scaled between a baseline $b_m$, the mean response of the panel, and a replicate anchor $a_m$, one half of the real cells scored against the other,
\begin{equation}
s=\tfrac16\bigl(s_{\mathrm{pds}}+s_{\mathrm{mse}}+s_{\mathrm{nmae}}+s_{\mathrm{fid}}+s_{\mathrm{reach}}+s_{\mathrm{jac}}\bigr),\qquad s_m=\frac{u_m-b_m}{a_m-b_m}.
\end{equation}
The baseline thus scores 0 and a replicate 1 by construction, and most of the score rests on per-gene means rather than on single cells. For one line, let $\mathcal P$ be the panel ($n=300$). The counts of a group of cells are summed and log-normalised once, giving the pseudobulk profile $\mu_p$ of the real cells of target $p$ and $\hat\mu_p$ of the predicted ones, with total $5\cdot10^4$, and $\mu_0$ for the controls. The first metric, \texttt{pds}, asks whether each predicted response is closest to the real response of its own target. With effects $\hat\delta_p=\hat\mu_p-\mu_0$ and $\delta_q=\mu_q-\mu_0$ taken on the genes outside the panel targets, and $d_{pq}=\mathrm{clip}_{[0,2]}(1-\cos(\hat\delta_p,\delta_q))$,
\begin{equation}
\mathrm{PDS}_p=1-\frac{\#\{q:d_{pq}<d_{pp}\}+\frac12(\#\{q:d_{pq}=d_{pp}\}-1)}{n-1}.
\end{equation}
The second, \texttt{mse}, is the error between predicted and real profiles corrected for sampling noise. With the delete-one jackknife noise $J_p$ of a real profile, $\hat J_p$ of a predicted one, and target weights $w_p$,
\begin{equation}
u_{\mathrm{mse}}=\frac{\sum_pw_p\bigl[\sum_{g}(\hat\mu_{pg}-\mu_{pg})^2-r\min(\hat J_p,J_p)-J_p\bigr]}{\sum_pw_p\bigl[\sum_{g}(\mu_{pg}-\mu_{0g})^2-J_p-J_0\bigr]},\quad r=\min\Bigl(1,\frac{B_{\mathrm{pred}}}{\mathrm{claim}}\Bigr),
\end{equation}
where $\mathrm{claim}=\sum_pw_p\min(\hat J_p,J_p)$ is the noise credit that a prediction may subtract and $B_{\mathrm{pred}}$ is the spread of the predictions between targets. A prediction of the control gives about 1, and writing $c$ for the cosine to the true shift and $\rho$ for the size ratio, $u_{\mathrm{mse}}\approx1-2\rho c+\rho^2$. The scaled metric is set to 0 whenever $u_{\mathrm{mse}}$ exceeds the baseline value of 0.989, which is why every early entry scored 0 on it. The remaining four metrics use the differentially expressed (DE) genes of each target, found by a Wilcoxon test against the real controls on genes above 5 CPM, BH-adjusted at 0.05, which gives the real set $R_p$ and the predicted set $\hat P_p$. Of these, \texttt{nmae} is the error of log$_2$ fold changes $\ell$ on the real DE genes, $\sum|\hat\ell-\ell|/\sum|\ell|$ over $R_p$, and \texttt{jac} is the overlap $|R_p\cap\hat P_p|/|R_p\cup\hat P_p|$. The metric \texttt{fid} counts sign agreements divided by $\max(|S_p|,|R_p|)$, where $S_p$ is the set of predicted DE genes with a defined real sign, and \texttt{reach} is the largest fraction of $R_p$ reached in the prediction's own ranking while the sign purity stays at or above 0.9. Our local scorers used the official baseline and replicate values of each metric as approximate anchors, namely 0.500 and 0.955 for \texttt{pds}, 0.989 and 0.036 for \texttt{mse}, 1.001 and 0.400 for \texttt{nmae}, 0.513 and 0.813 for \texttt{fid}, 0.072 and 0.968 for \texttt{reach}, and 0.029 and 0.399 for \texttt{jac}, so that a local score sits on the same scale as an official one.

\subsection{Leaderboards and Ranks}
\label{sec:testfield}

\begin{figure}[t]
\centering
\includegraphics[width=\linewidth]{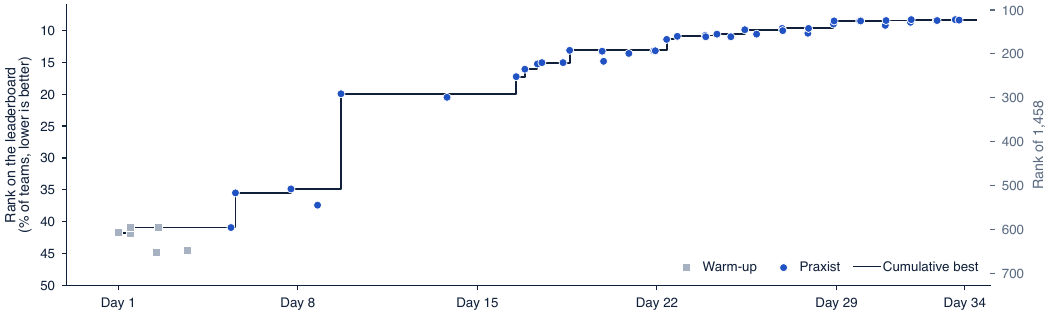}
\caption{\textbf{Rank over the campaign.} Each official submission placed among the final entries of the 1{,}458 teams on the validation leaderboard, with the cumulative best. Every point is ranked against this same field, however many teams had entered by its day. The left axis gives the percentage of teams and the right axis the rank.}
\label{fig:rank}
\end{figure}

Each team has two validation submissions per UTC day, and the public validation leaderboard keeps only a team's latest entry, so a submission made to test an idea can lower the visible rank. We date events by campaign day. Day 1 is 19 September 2026 and Day 34 is 22 October, when the validation leaderboard froze, and the first entry, scored just before midnight on 18 September, counts as Day 0. After the freeze, the organisers removed 177 teams without changing any score, 154 of them before the leaderboard was republished in November and 23 when the final results were published on 20 November, and they had restored one of them by 24 November. Every rank in this report places a score among the final entries of the 1{,}458 teams on that last version (\tab{tab:board}). We do not report live ranks, because the live leaderboard grew from about 1{,}040 to 1{,}630 teams over the campaign while its field improved as fast as we did, so a live rank mixes our progress with everyone else's. On the fixed field, the first entry ranks 608th (41.7\%) and the best entry, which scored 0.2344, ranks 122nd (8.4\%), and the largest rise came in a single step at CP3, from 508th to 291st (\fig{fig:rank}). The leaderboard itself lists our final entry, a 16-model ensemble of the best one, at 0.2342 and 123rd, because it keeps only the latest submission.

\begin{table}[t]
\centering\small
\caption{\textbf{Validation leaderboard as frozen on Day 34, 1{,}458 teams.} As published with the final results on 20 November and read on 24 November, after the organisers removed 177 teams and restored one, without changing any score. It keeps each team's latest entry. The six metrics are scaled, and the score is their mean.}
\label{tab:board}
\begin{tabular}{@{}rllrrrrrrr@{}}
\toprule
Rank & Team & Entry & Score & pds & mse & nmae & fid & reach & jac\\
\midrule
1 & Illumina AI & iCell & 0.4720 & 0.870 & 0.793 & 0.444 & 0.177 & 0.470 & 0.078\\
2 & X-team & X-99 & 0.3527 & 0.773 & 0.508 & 0.295 & 0.069 & 0.378 & 0.092\\
3 & CAS & CellMelody & 0.3447 & 0.850 & 0.531 & 0.295 & 0.085 & 0.289 & 0.019\\
\multicolumn{10}{c}{$\cdots$}\\
121 & Houlin Yu & celltics-1 & 0.2348 & 0.763 & 0.326 & 0.125 & 0.000 & 0.182 & 0.014\\
122 & peiyu li 2 & pl-18 & 0.2343 & 0.765 & 0.256 & 0.127 & 0.014 & 0.236 & 0.007\\
\rowcolor{rowhl}123 & \textbf{VenusCell} & t9\_x16 & 0.2342 & 0.784 & 0.296 & 0.103 & $-$0.003 & 0.226 & 0.000\\
124 & Ian Driver & geoatlas-v8.51 & 0.2333 & 0.746 & 0.256 & 0.149 & 0.017 & 0.209 & 0.022\\
125 & kai zhang2 & 26-kz-64 & 0.2320 & 0.763 & 0.185 & 0.174 & 0.028 & 0.239 & 0.003\\
\multicolumn{10}{c}{$\cdots$}\\
\bottomrule
\end{tabular}
\end{table}

\begin{table}[t]
\centering\small
\caption{\textbf{Final test leaderboard, 807 teams.} Scored on the blind lines D, E and F, published on 20 November and read on 24 November. It keeps each team's final entry. Our entry is \texttt{t9\_x16} with an exact per-cell realisation of its predicted changes, chosen after the freeze, which leaves the average predicted profile of each target unchanged.}
\label{tab:testboard}
\begin{tabular}{@{}rllrrrrrrr@{}}
\toprule
Rank & Team & Entry & Score & pds & mse & nmae & fid & reach & jac\\
\midrule
1 & Illumina AI & CellularAI & 0.4399 & 0.831 & 0.684 & 0.440 & 0.191 & 0.429 & 0.064\\
2 & FlowState & flowstate & 0.3348 & 0.836 & 0.363 & 0.329 & 0.169 & 0.313 & $-$0.002\\
3 & X-team & X-101 & 0.3263 & 0.735 & 0.468 & 0.304 & 0.066 & 0.319 & 0.067\\
\multicolumn{10}{c}{$\cdots$}\\
100 & CarlosVM & cvm-v27 & 0.2209 & 0.750 & 0.151 & 0.200 & 0.072 & 0.178 & $-$0.026\\
101 & The RL team & rl-20261025-final & 0.2208 & 0.673 & 0.174 & 0.186 & 0.073 & 0.176 & 0.043\\
\rowcolor{rowhl}102 & \textbf{VenusCell} & t9\_x16R & 0.2199 & 0.722 & 0.225 & 0.134 & 0.049 & 0.191 & $-$0.002\\
103 & JQ & zs-v21 & 0.2196 & 0.742 & 0.303 & 0.118 & $-$0.017 & 0.146 & 0.025\\
104 & Perfechoice & pc-f07-txf-a4 & 0.2195 & 0.737 & 0.182 & 0.162 & 0.060 & 0.165 & 0.010\\
\multicolumn{10}{c}{$\cdots$}\\
\bottomrule
\end{tabular}
\vspace{0.8em}
\centering\footnotesize
\caption{\textbf{Entry into the test phase by validation score.} Teams on the validation leaderboard as read on 24 November, and the share of them with an entry on the test leaderboard.}
\label{tab:entry}
\begin{tabular}{@{}lrrrrrr@{}}
\toprule
Validation score & Below 0 & 0 to 0.05 & 0.05 to 0.1 & 0.1 to 0.2 & 0.2 or above & All\\
\midrule
Teams & 540 & 126 & 169 & 411 & 212 & 1{,}458\\
Entered the test phase & 29.8\% & 39.7\% & 35.5\% & 49.9\% & 66.5\% & 42.3\%\\
\bottomrule
\end{tabular}
\end{table}

The final phase stayed blind until its leaderboard was published on 20 November, and it now holds 807 teams (\tab{tab:testboard}). Our test entry scored 0.2199, 102nd (12.6\%), and the two leaderboards hold different fields, a difference that changes how a rank should be read. Only 617 of the validation teams entered the test phase, and the share that entered rose broadly with the validation score, from 30\% of the teams with a negative score to 67\% of those at 0.2 or above (\tab{tab:entry}). The teams that stayed out had mostly stopped working on the challenge, since about half of them made their last validation submission before October, and they made a median of 6 submissions against 17 for the teams that entered. The other 190 test teams have no entry on the validation leaderboard, and most of them scored poorly, with a median of $-0.003$ and only 4 places in the top 100. Around our entry the test field was therefore stronger than the validation field, and this accounts for most of the change in our share of the leaderboard, from 8.4\% to 12.6\%. Measured within the same field, the change is smaller. Among the 617 teams on both leaderboards, \vc ranked 83rd (13.5\%) on validation and 98th (15.9\%) on test, consistent with a test score that fell by about as much as those of other strong entries.

\section{The Research Trajectory in Detail}
\label{sec:route}

This appendix supports the account of the trajectory in Sections~\ref{sec:researchtraj} and~\ref{sec:contrib}. It maps the runs of the campaign to its days, follows the route from the baseline to the best entry, shows how the official evaluations were spent, and measures how widely the search explored. The components named along the way are those of the final pipeline, which Appendix~\ref{sec:method} describes. The generative model named below is its flow-matching component, one of three learned parts, which corrects the direction of each response transferred from public screens.

\begin{table}[t]
\centering\footnotesize
\caption{\textbf{Phases of the campaign.} Runs are numbered launches, and runs 1--4 are the sessions of the warm-up. Local evaluations are counted by calendar day, so a run that crosses two phases is split.}
\label{tab:phases}
\begin{tabularx}{\linewidth}{@{}>{\raggedright\arraybackslash}p{3.6cm}llXr@{}}
\toprule
Phase & Days & Runs & Autoresearch harness and backend & Local evaluations\\
\midrule
Warm-up & 0--3 & 1--4 & AutoScientists, Claude Opus 5 & 121\\
First \px run & 4--5 & 5 & Open-source \px, DeepSeek & 70\\
Harness shakedown & 5--7 & 6--11 & \px with continuous research, DeepSeek & 458\\
Integrity and \texttt{mse} & 7--10 & 12--15 & \px, DeepSeek & 679\\
Plateau and model families & 10--17 & 16--21 & \px, DeepSeek & 552\\
\textit{hESC thinned} & 17--22 & 22--26 & \px, asynchronous from Day 21, DeepSeek & 712\\
GPT-6.1 & 22--26 & 27--32 & Asynchronous \px, GPT-6.1 & 1{,}182\\
Alternating backends & 26--34 & 33--35 and fallbacks & Asynchronous \px, GPT-6.1 and DeepSeek & 776\\
\midrule
All & 0--34 & & & 4{,}550\\
\bottomrule
\end{tabularx}
\end{table}

\tab{tab:phases} divides the campaign into phases by what the search was working on, and the run numbers used in the rest of this appendix can be read against it. The first phase that \px ran found the gain of CP2 within a day. The next two phases rebuilt the harness and the local proxy while the search assembled its components for size and direction, which gave CP3 on Day 9. The plateau that followed ended once the general harness had read the grader's source code, and the phases from Day 17 on ran on \textit{hESC thinned}, the version of the proxy matched to the depth of the challenge cells, and set most of the later checkpoints. The last phase spent more of its time on full retraining, which the fold-clean simulation of Appendix~\ref{app:scorers} made possible to judge.

\begin{figure}[t]
\centering
\includegraphics[width=\linewidth]{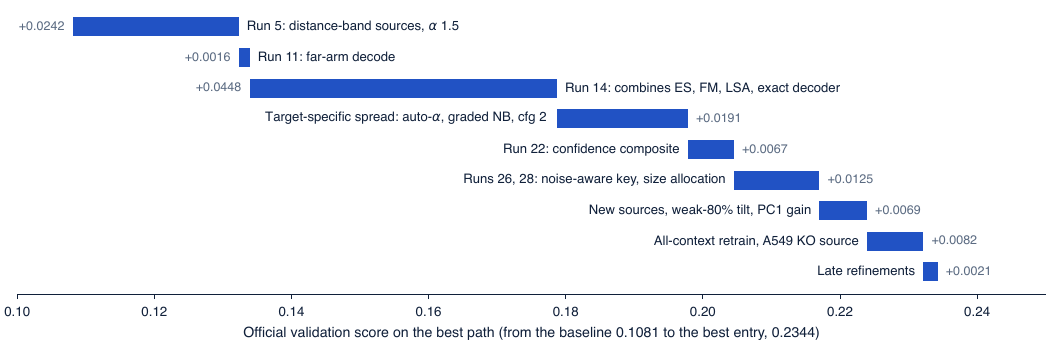}
\caption{\textbf{The route from the baseline to the best entry.} Steps of the cumulative best, grouped.}
\label{fig:waterfall}
\end{figure}

\subsection{The Route to the Best Entry}

From the baseline of 0.1081 to the best entry of 0.2344, the score rose by +0.1262 over 21 new bests after the baseline, and \fig{fig:waterfall} groups these steps into three phases of the work. The first phase was about which public lines to trust. Weighting the source lines by their distance to the target line in run 5 made each predicted response easier to tell apart from the others and ended the plateau of simple transfer. The second phase was about the size of each response. Run 14 brought together a model of response size, a flow-matching model of direction and learned attention over source lines with a decoder that keeps the predicted average of each target exact, and it made the largest single step of the month (+0.0448). Work on the target-specific spread, graded decoding and the guidance of the flow-matching model then added +0.0191. The third phase was about evidence and data. Weighting predictions by the evidence behind them in run 22, a ranking key corrected for noise in run 26 and a separate allocation of response sizes in run 28 together added +0.0192, and new source datasets, the removal of a line's own control programmes from weak targets and retraining on all public lines added a further +0.0172. Along this route the runs of the autoresearch harness set the steps of CP2, CP3, CP5 and CP6, and the late steps that relied on official feedback and full retraining were prepared alongside the search.

\subsection{Spending the Official Evaluations}
\label{sec:official}

With two results a day, every official evaluation was a decision, so we classified all 47 by their purpose at the time of submission, using the note recorded with each one (\tab{tab:spend}). Of these, 36 were submitted as expected improvements, 9 were probes that asked a question with no gain expected, and 2 repeated an earlier entry. The expected improvements set a new best 23 times, two of them in the warm-up, a hit rate of 64\%. One might expect this rate to fall once the standing rule of Day 21 made two submissions a day the norm, but it barely changed, at 10 of 15 before the rule and 13 of 21 after it. What changed was the size of the steps, since the 13 later bests added +0.0298 in total with a median of +0.0013, and the last four together added only +0.0021, each comparable to the seed noise of the generative model, whose standard deviation was 0.0004 to 0.0008. The misses were informative as well. Several losses of a candidate that the proxy preferred exposed a weakness of the proxy and led directly to a new version of it or to a lesson, as Appendix~\ref{app:scorers} records. The probes paid off less directly but twice decisively. The first official test of the model of response size on Day 8 scored below the best entry, yet it earned the first credit above 0 on \texttt{mse}, and that model of response size carried run 14 to the largest step of the campaign. On Day 23, a model retrained on all public lines was submitted as insurance for the blind phase and tied the best entry, which showed that training on the held-out lines did no harm and opened the way to the final entry. The other probes closed directions cheaply, among them an expert for one cell type on one line, a boost of the most confident genes, a transfer of control states that acts on one line only, more training contexts for the flow-matching model and larger ensembles. A rationed grader was therefore best spent on both expected gains and pointed questions, with every disagreement between proxy and grader fed back into the proxy.

\begin{table}[tb]
\centering\footnotesize
\caption{\textbf{Use of the 47 official evaluations.} Purpose comes from the note written with each submission, and a hit is a new official best. Probes asked a question with no gain expected.}
\label{tab:spend}
\begin{tabular}{@{}lrrrrr@{}}
\toprule
Period & Submissions & Expected gains & Hits & Probes & Repeats\\
\midrule
Days 0--20 & 21 & 15 & 10 & 4 & 2\\
Days 21--33 & 26 & 21 & 13 & 5 & 0\\
\midrule
All & 47 & 36 & 23 & 9 & 2\\
\bottomrule
\end{tabular}
\end{table}

\subsection{Breadth of the Search}
\label{sec:breadth}

All 40 new entries after the baseline belong to one line of descent, because each run started from the deployed entry, which raises the question of whether the search narrowed over time. We found instead that the range of ideas widened over the month. In a lineage record compiled on Day 31, where each of 265 key variants carries one family label assigned from the run reports, the 15 runs numbered 5 to 21 introduced 10 new families of related changes, whereas the 14 runs from run 22 to Day 28, where the record ends, introduced 37. Wider exploration did not change what survived. Latent and gene-space models of 110M and 190M parameters, the Stack foundation model and a deterministic residual head all lost to a pipeline that had been calibrated for weeks, and every kept step extended the previous best. An alternative pipeline had to beat a month of calibration within its first attempts and clear a fixed bar of +0.005 on the local proxy. A long campaign therefore favours its incumbent even when its exploration is wide, and a system meant to keep discovering should give new directions a protected period of tuning before they face the incumbent.

\section{Harness Improvement in Detail}
\label{sec:metaharness}

This appendix supports \secref{sec:harness}. It lists every change to the harness with the failure that prompted it, describes how knowledge was carried from one run to the next, and measures what the asynchronous update and the backend models changed in operation. \tab{tab:harness} groups the changes made by both sources of \secref{sec:harness}, the updates of \px from its developers and the changes of the general harness between them, by the kind of failure behind them. Most rows record a need that a run of two days would never reveal, such as storage that fails under concurrent load, an allowance that runs out mid-run or a scorer that a long search learns to exploit, and in a campaign of this length improving the harness meant repairing and extending the system while the research kept running.

\begin{table}[t]
\centering\footnotesize
\caption{\textbf{Changes to the harness and the failures behind them.} Run numbers can be read against \tab{tab:phases}.}
\label{tab:harness}
\begin{tabularx}{\linewidth}{@{}>{\raggedright\arraybackslash}p{1.9cm}>{\raggedright\arraybackslash}X>{\raggedright\arraybackslash}X@{}}
\toprule
Kind & What went wrong & Change and effect\\
\midrule
Scorer integrity & In run 12, a model trained on held-out labels scored near the replicate ceiling, and 11 later variants inherited it. Run 15 fitted weights to the held-out genes. In run 20, the best candidate used 1{,}000 cells per target, over the grader's limit. & Automatic checks of training manifests, of parameters fitted on held-out data, of seeds and of cell counts. The leaked runs were discarded.\\
\addlinespace[2pt]
Local proxy & \texttt{mse} stayed at 0 in run 5, and in run 6 changes that act only on lines near the sources passed the regression check. Later, the depth and the cell type of the held-out line misled the search. & Raw \texttt{mse} as an objective and official anchors as the primary score on Day 6, \textit{hESC thinned} on Day 17, the cancer set on Day 25 and the fold-clean simulation on Day 28 (Appendix~\ref{app:scorers}).\\
\addlinespace[2pt]
Throughput & Most GPUs sat idle on Day 6, and steps bound by the CPU held GPU slots on Day 25. & Multi-GPU profiles, a batch screening tool and a job queue. Scoring on the GPU made a complete evaluation 3.4$\times$ faster.\\
\addlinespace[2pt]
Reliability & Storage and API failures interrupted several runs. & More robust state handling, backups and automatic recovery, and updates of \px from its developers.\\
\addlinespace[2pt]
Continuity & Until Day 21, each new run waited for the researcher's approval. & The general harness launched each new run with its brief without waiting.\\
\addlinespace[2pt]
Backend and allowance & Reasoning effort was costly, and the weekly allowance ran out. & Effort lowered to high, with automatic fallback to DeepSeek and a switch back after each reset.\\
\addlinespace[2pt]
Knowledge & Each official result changed what the next run should know. & The brief of each run carried the deployed entry, a higher bar and the closed directions.\\
\bottomrule
\end{tabularx}
\end{table}

\subsection{Knowledge across Runs}
\label{sec:knowledge}

Knowledge had to survive every one of these changes, and both harnesses kept their own. The autoresearch harness recorded the evidence of its runs, while the general harness kept a report for every run and a numbered lessons file written for both people and agents, read the evidence record of \px whenever it needed to, and drew on both to write the brief of each new run. Over 26 revisions, the lessons file grew from 40 items on Day 5 to 62 by Day 30, and from about 2{,}000 to 4{,}050 words. It was almost append-only, with 10 lines deleted over the whole campaign, and two lessons were explicitly corrected, on Days 7 and 19, when an official result contradicted them. More often, later evidence qualified an earlier lesson in a new entry rather than rewriting it, so the file kept its history but grew harder to read, since a reader had to know which entry superseded which. Closed directions also returned in new forms, sometimes usefully, as when shrinkage toward zero for weakly supported changes, which cost 0.018 officially on Day 2, came back in run 22 as per-target evidence weights and became part of CP5. The upkeep of this knowledge, from deciding what a result meant to recording which lesson it replaced, was part of the work that people steered.

\subsection{Operating Efficiency and the Backend}
\label{sec:efficiency}

The asynchronous update of Day 21 aimed at operating efficiency, so we measured its effect from the job logs of every run (\tab{tab:efficiency}). A fair comparison holds the backend model fixed and counts like with like, because a job ranges from a few seconds to many hours, so we count full evaluations, each of which scores a candidate on the complete held-out sets, and compare DeepSeek V4 Pro with V4 Pro. The update kept the node busier and the campaign alive for longer. A run was live for 96\% of the 13 days after the update against 76\% of the 13 days before it, although part of this gain came from changes made in the same period, namely automatic recovery, backend fallback and the general harness launching each new run without waiting for approval. The share of run time with a GPU job rose from 91\% to 98\%, in part because one mostly idle run fell in the earlier window. The closest test of throughput is the five days around the update, when model and scorer were unchanged, and there runs 25--26 completed 4.7 full evaluations per run hour against 2.7 for runs 22--24, although these runs total only 23 hours. The largest change in the use of GPUs came earlier, when evaluation moved onto them and raised their use from 0.17 GPU-hours per run hour in the open-source run to 4.6 in steady continuous operation. These comparisons are observational, since the job mix, the artifact and the stage of the research all changed along the way, but read together they suggest that the update brought continuity and fuller use of the node, while throughput depended at least as much on the backend and on the work in hand.

\begin{table}[t]
\centering\footnotesize
\caption{\textbf{Operating efficiency by stage and backend model, from the job logs.} Continuous covers the steady runs 14--24. A full evaluation scores one candidate on the complete held-out sets.}
\label{tab:efficiency}
\begin{tabular}{@{}lrrrrr@{}}
\toprule
& Open source & Continuous & \multicolumn{3}{c}{Asynchronous}\\
\cmidrule(lr){4-6}
Backend model & V4 Pro & V4 Pro & V4 Pro & V4.1 Flash & GPT-6.1\\
\midrule
Run hours & 12.7 & 233.4 & 142.3 & 50.5 & 102.4\\
GPU-hours per run hour & 0.17 & 4.62 & 4.82 & 4.41 & 0.86\\
Share of run time with a GPU job (\%) & 13.3 & 90.6 & 98.3 & 94.3 & 32.6\\
Mean job length (minutes) & 4.8 & 14.6 & 26.0 & 9.0 & 3.1\\
Full evaluations per run hour & 8.50 & 2.46 & 1.43 & 6.20 & 0.63\\
\bottomrule
\end{tabular}
\end{table}

The backend models differed far more from one another than the harness versions did. DeepSeek V4.1 Flash reached 6.2 full evaluations per run hour, whereas GPT-6.1 reached 0.6. Their effect on the score, however, is confounded with nearly everything else, because the backend changed 8 times in step with the harness version, the stage of the research and the allowance of each provider. A few observations nevertheless bound it. Of the runs that produced a new best directly, five ran on V4 Pro (runs 5, 11, 14, 22 and 26) and one on GPT-6.1 (run 28), which found its size allocation within about two hours before a storage failure stalled it. The later GPT-6.1 runs were mostly cut short by the allowance or by storage failures. Switching backends therefore made no difference to the score that this campaign can detect, and a matched comparison of backends on the same artifact and harness remains open.

\section{The Local Proxy in Detail}
\label{app:scorers}

This appendix supports \secref{sec:proxies}. It defines every version of the local proxy, records why each one replaced or joined the last, and explains how \fig{fig:proxy} places them against the official grader. Every version held out a set of public lines and targets from training, predicted them as new lines, and scored the predictions with the six metrics of the official grader (Appendix~\ref{app:score}). The versions therefore differ from the grader, and from one another, mainly in which lines and targets they hold out and in the sequencing depth of the held-out cells (\tab{tab:scorers}). Each change answered a failure of the version before it, revealed by an official result or by the search itself, except the last, which answered the change of goal to the blind test, and \tab{tab:scorerchange} gives the reason and the evidence for each version. Besides the four versions of \secref{sec:proxies}, a cancer set served as a further check for a few days.

\begin{table}[tb]
\centering\footnotesize
\caption{\textbf{Versions of the local proxy.} From is the first day it chose candidates. Depth is the median count of RNA molecules (UMI) per held-out cell. The first three report a score, and the last two a change against a reference entry.}
\label{tab:scorers}
\begin{tabularx}{\linewidth}{@{}llXXl@{}}
\toprule
Version & From & Held-out lines & Targets & Depth\\
\midrule
\textit{Nadig} & Day 0 & HepG2, Jurkat & 100, then 150 per line & Native\\
\textit{hESC} & Day 2 & H1 hESC & 150 of VCC 2025 & 54k\\
\textit{hESC thinned} & Day 17 & H1 hESC & 150 of VCC 2025 & 20k\\
Cancer set & Day 25 & 6 cancer and normal lines & Up to 200 per line & Up to 20k\\
\textit{Fold-clean simulation} & Day 28 & 9 cancer, normal and stem lines & 3 sets of 300 per line & Up to 20k\\
\bottomrule
\end{tabularx}
\end{table}

The first three versions share their construction. \textit{Nadig} holds out the HepG2 and Jurkat screens of \citet{nadig2025}, first with 100 targets per line and then with 150 targets of at least 80 cells, which brought its own anchors close to the official ones. Because all its targets are essential genes with strong effects, it reads higher than the grader. From Day 6 it served only as a regression check, on which a candidate had to stay within 0.01 of the deployed model, since its lines lie near the sources and a gain on them can come from changes that never act on the challenge lines. \textit{hESC} holds out the H1 line of the 2025 challenge \citep{roohani2025vcc} from the sources and from every learned part, and scores its 150 targets at 400 cells each, first with the anchors of its own panel and from Day 2 with the official anchors, which became the primary score of \px on Day 6. \textit{hESC thinned} is the same held-out set with the real and control cells binomially thinned from 54k to the official median of 20k UMI per cell. At full depth the line sat in the wrong regime for three mechanisms of the grader defined in Appendix~\ref{app:score}, the noise credit of \texttt{mse}, the number of differentially expressed genes per target and the ordering behind \texttt{reach}, and thinning repaired all three. Candidates on it were scored over two to five seeds, and it stayed the scorer of every run of \px up to the freeze.

\begin{table}[t]
\centering\footnotesize
\caption{\textbf{Why each version of the local proxy changed.} Day is the first day of selection, as in \tab{tab:scorers}, support is the evidence for the change, and run numbers can be read against \tab{tab:phases}.}
\label{tab:scorerchange}
\begin{tabularx}{\linewidth}{@{}r>{\raggedright\arraybackslash}p{3.4cm}>{\raggedright\arraybackslash}X>{\raggedright\arraybackslash}X@{}}
\toprule
Day & Change & Why & Support\\
\midrule
2 & \textit{Nadig} to \textit{hESC} & \textit{Nadig} read 0.33 for an entry that scored 0.10. All its targets are strong essential genes, and its lines lie close to the sources. & \textit{hESC} read 0.1076 for the deployed entry, which scored 0.1081 officially.\\
\addlinespace[2pt]
2 & Official anchors reported alongside, and a candidate must beat the deployed model on both versions & A shrinkage entry read 0.209 and lost 0.018 officially, because the baseline of \textit{hESC} makes almost no DE calls and so credited any full set of predicted calls. & The order stayed wrong on \textit{hESC}, and only \textit{Nadig} flagged the loss.\\
\addlinespace[2pt]
6 & Raw \texttt{mse} as an objective, primary score on official anchors, \textit{Nadig} as a regression check & Scaled \texttt{mse} was 0 in 108 of 108 rows of run 5, which gained +0.1554 locally for +0.0242 officially. In run 6, changes acting only on near lines passed \textit{Nadig}. & In run 14, 113 of 158 complete evaluations read raw \texttt{mse} below the threshold of 0.989.\\
\addlinespace[2pt]
17 & \textit{hESC thinned} & Official gains read as local losses, and at 54k UMI the noise credit, the DE genes and the order of \texttt{reach} all behaved unlike the challenge lines. & Four official entries ranked in their official order, from 0.2303 to 0.2527 locally and 0.1787 to 0.1964 officially.\\
\addlinespace[2pt]
25 & Cancer set added & A new best read as a tie on \textit{hESC thinned}, and a loss read as a gain, pointing to the stem-cell state of hESC. & It gave the right sign for 4 of 4 official changes on a test fixed in advance, against 1 of 3 for \textit{hESC thinned}.\\
\addlinespace[2pt]
28 & \textit{Fold-clean simulation} & The final entry would meet new lines blind, the leading model had trained on \textit{hESC} and \textit{Nadig}, and a loss on the cancer set came from a reference model that had seen the line. & With clean models that loss became a gain, and the entry it supported set a new best of +0.0061.\\
\addlinespace[2pt]
28 & Calibrated change in place of raw change & The cancer set read +0.0192 for a change that lost 0.0033 officially, since local and official metrics move by different amounts. & Right sign on both pairs of the fit, although each metric weight rests on two or three pairs.\\
\bottomrule
\end{tabularx}
\end{table}

The last two versions answered different needs. The cancer set holds out six lines, HCT116, RPE1, TeloHAEC, A549 under CRISPRi, DLD-1 and A549 under knockout, and it forces the branch of the model meant for distant lines, as on the challenge lines, since these lines share their measurement chemistry with the sources and so look nearer to them than the challenge lines do. Each of its sets puts the official targets first and adds up to 200 targets covered by at least two sources, with up to 400 cells per target and 4{,}000 controls, thinned to 20k UMI where a line is deeper, and it reports the mean change against a reference over two or three seeds. The \textit{fold-clean simulation} instead imitates the blind final phase. Each of nine lines, the six above with K562, HEK293T and KOLF, becomes a new line in turn, removed with its related contexts from the sources and from every learned part. Each line has three sets of 300 targets that differ in how well the sources cover them, one matched to the official panel, one thinly covered and one drawn at random, with two replicates each, so a candidate is scored on up to 54 combinations of line, set and replicate. K562, HEK293T and KOLF were never used to choose a candidate and stand in for the blind test lines. The simulation screened the leading candidates before submission and never replaced \textit{hESC thinned}. Its first screens reported the raw change, and later ones reported a calibrated change $\sum_m\beta_m\Delta_m/6$, with metric weights $\beta_m$ fitted on official pairs, to correct for metrics that move more locally than they do on the grader.

\begin{table}[t]
\centering\footnotesize
\caption{\textbf{Local and official changes of selected entries.} Each change is against the entry it extended.}
\label{tab:proxy}
\begin{tabular}{@{}rlrrl@{}}
\toprule
Day & Entry & Local $\Delta$ & Official $\Delta$ & Reading\\
\midrule
5 & Source weighting by similarity (CP2) & +0.1554 & +0.0242 & right sign, 6$\times$ too large\\
7 & Next new best of \px & +0.0106 & +0.0016 & right sign, 6$\times$ too large\\
9 & Size and direction combined (CP3) & +0.04 & +0.0448 & about 1:1\\
16 & Target-specific part enlarged & $-$0.010 & +0.0115 & wrong sign, fixed by thinning\\
21 & Per-target blend of the flow model & +0.0029 & $-$0.0002 & winner's curse\\
24 & Weak targets tilted, sizes locked & +0.0032 & $-$0.0008 & wrong sign\\
27 & Gain on one line's control programme & +0.0192 & $-$0.0033 & blind to the one line it changed\\
30 & Blend with an external model & +0.0006 & $-$0.0054 & wrong sign\\
31 & Deterministic head in place of the flow model & +0.0027 & $-$0.0031 & wrong sign\\
\bottomrule
\end{tabular}
\end{table}

As official results accumulated, the relation between local and official changes moved through three regimes (\tab{tab:proxy}). Early on, official gains were about six times smaller than local ones, in the middle of the campaign the two roughly matched, and for some late changes they had opposite signs. Winners also carried selection bias, since after about 100 variants on the same held-out set a local gain of +0.003 did not transfer, a winner's curse that became one of the lessons of the campaign. The checks on the integrity of local scores mattered as much as the repairs of the proxy, because without them the search would have chased leaked or censored scores (\tab{tab:harness}). \fig{fig:proxy}a needs two further notes. Four entries have only a logged change on \textit{hESC thinned}, and each takes the score of its reference plus that change. For the \textit{fold-clean simulation}, each of the twelve submitted entries that it screened is shown at its mean over the 54 combinations that the screens share, and the two entries screened on fewer combinations take their parent's mean plus the paired difference on the combinations they were screened on.

\section{Human Steering in Detail}
\label{app:coding}

This appendix supports \secref{sec:human}. It describes how the researcher's messages were coded, gives the themes behind the finding that rules held while judgments had to be repeated, and defines how unattended time was measured. Messages were taken from the session logs of the general harness, including messages queued while the agent was busy, from Day 0 to the freeze of the leaderboard. They were deduplicated within a day, automated prompts were removed, and each message received one class by its main function in a single pass of coding by hand (\tab{tab:codebook}). We ordered the classes by how much a person says, from a status check to steering in conversation \citep{agent-laboratory,tellonce}, with harness changes as a separate class at the level of the meta-harness. Of the 302 typed messages, 124 concerned administration or reporting, and the other 178 steered the campaign in only 2{,}997 words, spread over 34 of 35 days. The volume of messages did not track the score. Steering came in two bursts, on Days 4--6 around the start of \px and on Days 20--22 around the update of Day 21, which together hold 44\% of the steering messages, whereas Days 16--18 added +0.0259 on only 7 messages.

\begin{table}[t]
\centering\footnotesize
\caption{\textbf{Codebook of the researcher's messages, from Day 0 to the freeze of the leaderboard.} Level orders the steering classes by how much a person says. Words are not counted for the two classes that do not steer.}
\label{tab:codebook}
\begin{tabularx}{\linewidth}{@{}lcrrX@{}}
\toprule
Class & Level & Messages & Words & Definition\\
\midrule
Status check & 0 & 34 & 193 & Asks for progress and changes nothing.\\
Go-ahead & 1 & 21 & 77 & Lets the current plan continue, such as ``continue'' or ``good''.\\
Submission decision & 2 & 32 & 568 & Approves a submission or sets a rule for submitting.\\
Correction or rule & 3 & 20 & 326 & Fixes a behaviour that should hold from now on.\\
Research direction & 4 & 29 & 1{,}165 & Says what to try, read or investigate.\\
Harness change & meta & 42 & 668 & Changes or diagnoses \px, its backend or its launch policy.\\
Reporting & -- & 41 & -- & Asks for slides, write-ups or explanations.\\
Other & -- & 83 & -- & Data transfers, access setup, hand-over and other tasks.\\
\bottomrule
\end{tabularx}
\end{table}

\subsection{Inputs That Held and Inputs That Recurred}

To find which inputs stayed in the system once given, we grouped the steering messages into themes by keyword rules and checked every match by hand (\tab{tab:themes}). A theme counts as held when it was not restated after the rule or tool that addressed it existed. Six operational rules held after a single statement, among them to submit nothing unrequested, to stop a run when an allowance ends, to keep run state on network storage and never to resubmit an entry to restore a rank. Each could be written down and checked. The judgments about how to spend compute, effort and the grader did not hold, even when a tool was built for them, and even the observation that \texttt{mse} stayed at 0 took four messages over three days before an entry moved the metric. Consistent with this, steering per unit of progress rose over the month. The researcher sent 7.8 steering messages per day over Days 0--10, 3.2 over Days 11--21 and 4.4 over Days 22--34, which amounts to about 11.5, 13.5 and 19.1 messages per +0.01 of the best official score over each period.

\begin{table}[t]
\centering\footnotesize
\caption{\textbf{Themes of steering and whether they held.} A theme held when the researcher did not restate it once the rule or tool that addressed it was in place, which the table records for each theme.}
\label{tab:themes}
\begin{tabularx}{\linewidth}{@{}>{\raggedright\arraybackslash}p{4.8cm}rrlX@{}}
\toprule
Theme & Messages & Days & First to last & Held?\\
\midrule
Six operational rules & 1--2 each & 1 each & -- & Yes, each given on a single day\\
Use the idle GPUs & 14 & 7 & Days 6--33 & No, restated 4 times after the job queue of Day 27\\
Keep exploring after a local win & about 13 & about 11 & Days 8--33 & No\\
Read the literature while runs wait & 8 & 6 & Days 12--31 & No\\
Is anything worth submitting? & 11 & 9 & Days 5--33 & Partly, 5 times after the rule of Day 21\\
\texttt{mse} is always 0 & 4 & 3 & Days 5--8 & Moved by an entry after the fourth message\\
\bottomrule
\end{tabularx}
\end{table}

\subsection{Unattended Time}
\label{sec:unattended}

The 34 days of the campaign are calendar time with a person in the loop, so we measured how long the system worked without one. Run windows come from the run records, and a gap is the time between consecutive typed messages to the general harness, with automated prompts excluded. The live run time inside a gap is the overlap of the gap with any run window. A gap without a message is not a gap without intervention, since the general harness also stopped and relaunched runs on its own, so these figures measure human attendance. The longest stretch of live run time without a human message, about 21 hours from early on Day 17 to early on Day 18, fell inside run 22, the first run on \textit{hESC thinned}, so a single scorer judged every candidate in it. In that time, eight peers logged 259 evaluations, 85 of them complete evaluations over two or three seeds, and the integrity checks rejected 5. The peers stacked test-time adaptation of the flow-matching model, a depth-dependent count spread, evidence-weighted shrinkage and a new ranking key into one composite, and on that scorer the composite beat the deployed entry by +0.015 and +0.019 on two seeds. Two candidates from the stretch were submitted the next day, and while an early variant lost 0.0005, the composite set the new best of CP5 (+0.0067). No longer stretch of live run time passed without a person checking in.

\section{Cost in Detail}
\label{app:cost}

This appendix supports \secref{sec:cost}. It breaks the total of \totalcostv\ into its parts (\tab{tab:budget}) and says how each part was measured, so that the estimate can be checked or redone at other prices. Token usage was counted from the records of each harness. DeepSeek usage is summed from the local transcripts of the agents, deduplicated by message and request and bucketed by hour. Codex usage is the final cumulative total of each session in the run logs, spread evenly over the run for plotting, and Claude usage is summed from the records of the general harness. The DeepSeek tokens are priced at list price, at \$0.044 for a cache hit, \$1.32 for a miss and \$3.96 per million output tokens at peak for V4 Pro, and \$0.006, \$0.30 and \$1.20 for V4.1 Flash, both at half price off-peak. V4 Pro served until Day 32 and V4.1 Flash after that. The Pro price was read from DeepSeek's page on Day 6, and the Flash price comes from secondary sources. The two subscriptions are counted as one monthly fee each, Claude Ultra at \$200 for the general harness from Day 0 and a ChatGPT Team Premium seat at \$125 for Codex from Day 22.

\begin{table}[tb]
\centering\footnotesize
\caption{\textbf{Budget used up to the freeze of the leaderboard.} Tokens include cache reads, and dollars are estimates at list price, each rounded separately.}
\label{tab:budget}
\begin{tabular}{@{}lrr@{}}
\toprule
Tokens (billions) & All & Uncached\\
\midrule
DeepSeek API (\px) & 75.1 & 1.41\\
GPT-6.1 through Codex (\px) & 5.2 & 0.55\\
Claude Opus 5 (general harness) & 7.2 & 0.82\\
\midrule
GPU-hours used & \multicolumn{2}{r}{\gpuhv}\\
GPU cost at \$2.00 per GPU-hour & \multicolumn{2}{r}{\gpucostv}\\
DeepSeek V4 Pro at list price & \multicolumn{2}{r}{\dsprov}\\
DeepSeek V4.1 Flash at list price & \multicolumn{2}{r}{\dsflashv}\\
Claude Ultra subscription, one month & \multicolumn{2}{r}{\$200}\\
ChatGPT Team Premium seat, one month & \multicolumn{2}{r}{\$125}\\
\textbf{Total} & \multicolumn{2}{r}{\textbf{\totalcostv}}\\
\bottomrule
\end{tabular}
\end{table}

GPU time is counted as the time actually used rather than as the time the node was held. The node held 8 A100 GPUs and was dedicated to the campaign, but the work could equally have run from a shared job queue. Up to Day 27, GPU time is the run time of each job multiplied by the GPUs allocated to it, from the scheduler log of each \px run. From Day 27, a per-minute log of utilisation covers all work on the node, with a mean of 3.0 of the 8 GPUs active at 10\% utilisation or more. Training jobs that the general harness ran itself before Day 27 have no central log and are not included, and applying the measured 3.0 GPUs to the whole period gives 2{,}146 GPU-hours, about 6\% above the \gpuhv\ reported. GPU use in the warm-up, which ran on a shared cluster, was not recorded and is excluded. The price of \$2.00 per GPU-hour is an assumption, close to the median on-demand price of this GPU in 2026.

\section{VenusCell}
\label{sec:method}

This appendix describes \vc, the final method of the campaign and the pipeline whose components \secref{sec:researchtraj} follows as they entered. Rather than a single trained model, it is a pipeline that predicts a log-ratio response per target from public screens and the query line's own controls, and decodes it into counts (\fig{fig:pipeline}). No perturbed cell of a query line from the challenge enters at any point, although one validation line resembles a T-cell leukaemia line whose public screen is among the sources (Appendix~\ref{app:data}). Only three components are learned, each with contexts held out, and everything else is computed in closed form from the public screens and the line's controls.

\textbf{Sources.} For a source context $s$ with mean CP10k profiles $M_{st}$ of target $t$ and $c_s$ of controls, the stored response is the log ratio $L_{stg}=\log\bigl((M_{stg}+\epsilon)/(c_{sg}+\epsilon)\bigr)$, with $\epsilon=0.05$. Sources enter through coverage weights $\gamma_{st}\propto n_{st}/(n_{st}+20)$, where $n_{st}$ is the cell count. For a query line, the panel draws on about 35 pseudobulk tables, and the learned parts are trained on 34--37 contexts of the public screens in \tab{tab:datasets}.

\begin{figure}[!tb]
\centering
\resizebox{\linewidth}{!}{\begin{tikzpicture}[x=1cm,y=1cm,
  st/.style={rounded corners=2.5pt,draw=#1,line width=.6pt,fill=#1!6,align=center,font=\sffamily\fontsize{6.4}{7.6}\selectfont,inner sep=3pt,minimum height=1.15cm,text width=2.62cm},
  hd/.style={font=\sffamily\bfseries\fontsize{6.8}{8}\selectfont,text=chuman},
  ar/.style={-{Latex[length=3pt,width=3.8pt]},line width=.7pt,color=chuman}]
\node[st=cgray] (i) at (0,0) {35 source screens\\control cells of the line\\300 target genes};
\node[st=cpraxist] (d) at (3.25,0) {Coverage + LSA\\FM correction\\blend};
\node[st=coperator] (m) at (6.5,0) {Effect size, confidence\\auto-$\alpha$ spread\\size allocation, tilt, PC1 gain};
\node[st=cgreen] (c) at (9.75,0) {Rescaled real controls\\graded NB noise\\exact column sums};
\node[st=cgray] (o) at (13,0) {400 cells per target\\per line};
\node[hd] at (0,.82) {Inputs}; \node[hd] at (3.25,.82) {Direction}; \node[hd] at (6.5,.82) {Magnitude}; \node[hd] at (9.75,.82) {Decoder}; \node[hd] at (13,.82) {Output};
\draw[ar] (i)--(d); \draw[ar] (d)--(m); \draw[ar] (m)--(c); \draw[ar] (c)--(o);
\end{tikzpicture}}
\caption{\textbf{The \vc pipeline.}}
\label{fig:pipeline}
\end{figure}

\textbf{Direction.} The direction combines three estimates. The first, the coverage direction, is the weighted mean of $L_{st}$ over sources. The second, learned source attention (LSA), is an MLP over eight features of each (source, target) pair, such as similarity, cell counts and leave-one-out agreement, and it outputs softmax weights over sources. The third is a 256-dimensional latent conditional flow-matching (FM) model \cite{lipman2023fm} over cell sets. It is trained with optimal-transport pairing \cite{tong2024otcfm}, conditioned on the line's controls, a protein embedding of the target \cite{esm2} and the transfer prior, and sampled in 16 Euler steps with guidance weight 2 \cite{ho2021cfg}. At test time its set encoder adapts with 400 unsupervised steps on the line's controls, and only the model's shift beyond the prior is kept. The blend is $0.8\cdot\mathrm{eq}(\mathrm{LSA}+0.5\,\mathrm{FM})+0.2\cdot\mathrm{eq}(\mathrm{coverage})$, rescaled to the coverage norm, where the operator eq equalises the norms of its terms.

\textbf{Magnitude.} A small effect-size network sets the size of each target's response. Within a target, the 100 genes with the largest noise-corrected $|z|$-score keep full size, while the others shrink to 0.4. The target-specific part $S_t$ of the response, its deviation from the panel mean, is then scaled:
\begin{equation}
l_t=\bar g+a\,\varpi_t S_t,\qquad a\ \text{such that}\ \ \mathcal B(a)\approx 0.97^2\,\mathrm{claim},
\end{equation}
where $\mathcal B$ is the between-target spread of predicted profiles, claim is the sampling noise the scorer lets a prediction subtract, and $\varpi_t$ is a reliability weight from source coverage. This step is label-free and is computed per line. Three smaller steps follow. First, per-target size comes from a leave-one-cluster-out projection. Second, the 10 leading control principal components are removed from the weakest 80\% of targets at strength 0.75, with their energy kept. Finally, the line's own leading control axis gets a gain of 2 on the top 20\% of targets, behind a pluripotency gate.

\textbf{Decoder.} Each predicted cell is a real control cell of the line, rescaled so that the mean of the 400 cells equals $\exp(l_t)\odot\kappa$, where $\kappa$ is the mean raw count of the controls. We then add graded negative-binomial noise, with dispersion set by predicted confidence, and repair the column sums exactly. Because the score reads means and ranks, the decoder's job is to place these correctly rather than to model individual cells.

\textbf{Targets with few sources.} The direction of a target is read from the public screens that knocked it down, so a target that no screen covers has no response to transfer. For a target with fewer than two sources, the pipeline turns to its 20 nearest neighbours in a graph of co-dependent genes from DepMap \citep{tsherniak2017,depmap24q4}. A target with no source gets the panel's shared response plus a target-specific part along the mean response of these neighbours, and for a target with one source this direction is blended with that of its source. The rule mattered little, because every validation target had at least two sources and only one test target had fewer, with a single source. Such thinly covered targets are mostly not expressed in a held-out line, so their response is close to none, and on held-out lines the richer gene graphs of Appendix~\ref{app:data} scored lower than this rule.

\textbf{Origins.} \tab{tab:origin} lists the components and what each added. The ideas drew widely on the single-cell literature, of which the system read about 100 papers and wrote more than 40 reviews. For example, findings that simple baselines are hard to beat \citep{ahlmann2025} favoured transfer over large end-to-end models, and work on systematic variation \citep{systema} led to treating the shared response of a panel separately from the target-specific part. The flow component builds on flow matching with optimal-transport pairing \citep{lipman2023fm,tong2024otcfm} and on cell-population models \citep{cellot,cellflow}, and its conditioning on a set of controls follows State \citep{state2025}. The decoder follows count models for RNA-seq \citep{deseq2}, and the noise-corrected statistics use empirical-Bayes shrinkage \citep{ash}.

\begin{table}[t]
\centering\footnotesize
\caption{\textbf{Components of \vc.} Each component with the day it entered the best entry and the change in the official score at that step. Two steps of less than +0.002, on Days 7 and 23, are left out.}
\label{tab:origin}
\begin{tabularx}{\linewidth}{@{}Xlr@{}}
\toprule
Component & Entered & Official $\Delta$\\
\midrule
Log-ratio source transfer & Day 1 & --\\
Source lines weighted by similarity to the target line & Day 5 & +0.0242\\
Coverage weights, LSA, FM correction, effect size, exact NB decoder & Day 9 & +0.0448\\
Control-calibrated spread (auto-$\alpha$) & Day 16 & +0.0147\\
Graded NB, guidance weight 2 & Day 17 & +0.0044\\
Confidence top-100, evidence weights & Day 18 & +0.0067\\
Noise-aware ranking key & Day 22 & +0.0077\\
Cluster-projection size allocation & Day 22 & +0.0048\\
Pfizer and other new sources & Day 24 & +0.0019\\
Weak-80\% tilt & Days 25--26 & +0.0040\\
All-context retrain, A549 knockout source & Day 28 & +0.0082\\
Own-PC1 gain, FM on 37 contexts, pluripotency gate & Days 27--31 & +0.0025\\
\bottomrule
\end{tabularx}
\end{table}

\textbf{What the method shows.} \vc identifies which response belongs to which target far above chance on validation lines that it never saw perturbed. Its scaled \texttt{pds}, which checks that each prediction lies closest to its own target, is 0.78, above about 84\% of the top 400 teams. We found that most of the gain came from data and calibration rather than model size. First, at least 60\% of the official gain up to Day 24 came from changes that train no new model. Second, about 60\% of the largest late step came from deep CRISPRi screens used as sources. Finally, a 110M latent model and a 190M gene-space model both lost to the 27M latent model. For this task, cross-line averaging with careful calibration therefore proved hard to beat, in line with earlier benchmarks of simple baselines, and this is itself a useful scientific result. One calibration step, worth about +0.015, sets the size of the target-specific part of each response from the line's own controls. It needs no labels and corrects a known weakness of models that capture mostly the shared response \citep{systema}. In contrast, a few components have a biological reading, such as the gain on a line's leading control programme, which in one case resembles a p53-type arrest response and is switched off for pluripotent lines. The main open problem is the direction of the response. On held-out lines, the noise-corrected cosine similarity between predicted and true responses is 0.21 overall and 0.32 on strong targets. Moreover, about 47\% of the reproducible missing direction lies outside every subspace that sources or controls can reach. Further progress will therefore need new information about the query line, and not only better ways to combine what is public.

\section{Public Datasets}
\label{app:data}

This appendix lists every public perturbation dataset that the campaign used or tried (\tab{tab:datasets}), grouped by what it was used for. We supplied the first seven, and the agents found the rest, of which six entered the model or its flow-matching part and five were tried and dropped. Four further sets were used only after the freeze, as held-out lines on which to choose the blind test entry, and no component was trained on them. The two MOLM-13 screens among them were also tried as sources and dropped. The fallback of \vc for thinly covered targets (Appendix~\ref{sec:method}) takes a graph of co-dependent genes from DepMap \citep{tsherniak2017,depmap24q4}. DepMap expression and gene-effect data \citep{depmap24q4} were also tested as a prior on the size of each response and dropped, as were neighbour graphs from STRING \citep{string2023}, CORUM \citep{corum2023} and hu.MAP 3.0 \citep{humap3}. The MSigDB Hallmark gene sets \citep{liberzon2015} were used to describe the control programmes of each line. Before the campaign, the team inferred from the control cells of the three validation lines that one resembles a T-cell leukaemia line, which has a public screen among those we supplied, whereas the other two resemble cancer lines for which the team found no public Perturb-seq screen.

\begin{table}[t]
\centering\footnotesize
\caption{\textbf{Public perturbation datasets.} Each context is one line or condition. L1000 holds bulk signatures and the Jurkat readout covers about 400 genes, so neither has a comparable cell count. The two sets marked $^\dagger$ were taken from the scPerturb collection \citep{peidli2024}.}
\label{tab:datasets}
\begin{tabularx}{\linewidth}{@{}>{\raggedright\arraybackslash}p{6.3cm}>{\raggedright\arraybackslash}Xlrr@{}}
\toprule
Dataset & Cell line or type & Perturbation & Contexts & Cells (M)\\
\midrule
\multicolumn{5}{@{}l}{\textit{Supplied by the team and used in the model}}\\
X-Atlas/Orion \citep{orion2025} & HCT116, HEK293T & CRISPRi & 2 & 7.68\\
KOLF2.1J atlas \citep{nourreddine2026} & KOLF2.1J iPSC & CRISPRi & 1 & 2.58\\
Genome-scale Perturb-seq \citep{replogle2022} & K562, RPE1 & CRISPRi & 3 & 2.43\\
hPSC CRISPRi map \citep{feng2026} & Human iPSC lines & CRISPRi & 19 & 0.66\\
Essential-gene screens \citep{nadig2025} & HepG2, Jurkat & CRISPRi & 2 & 0.37\\
VCC 2025 training set \citep{roohani2025vcc} & H1 hESC & CRISPRi & 1 & 0.22\\
CPA-Perturb-seq \citep{kowalski2024} & HEK293T, K562 & CRISPRi & 2 & 0.07\\
\addlinespace[2pt]
\multicolumn{5}{@{}l}{\textit{Found by the agents and used in the model}}\\
H1 screen \citep{pan2026} & H1 hESC & CRISPRi & 1 & 0.93\\
Pfizer screen \citep{wong2026} & TeloHAEC, basal and TNF/IL-1$\beta$ & CRISPRi & 2 & 0.85\\
Myllia A549 \citep{fellinger2026} & A549 & CRISPRi & 1 & 0.05\\
\addlinespace[2pt]
\multicolumn{5}{@{}l}{\textit{Found by the agents and used in the flow-matching model only}}\\
PerturbMatch \citep{yeung2026} & DLD-1 & CRISPRi & 1 & 0.71\\
Biohub atlas \citep{liu2026atlas} & A549 & Knockout & 1 & 0.61\\
Endothelial screen \citep{schnitzler2024} & TeloHAEC & CRISPRi & 1 & 0.21\\
\addlinespace[2pt]
\multicolumn{5}{@{}l}{\textit{Found by the agents, tested and dropped}}\\
CD4$^{+}$ T-cell screen \citep{zhu2026} & Primary CD4$^{+}$ T cells & CRISPRi & 3 & 21.0\\
Pathway screens \citep{jiang2025} & A549, K562 and four other lines & CRISPRi & 30 & 1.64\\
DepMap Perturb-seq \citep{maffa2026} & 16 cancer lines & Knockout & 16 & 0.19\\
Jurkat TAP-seq \citep{song2025} & Jurkat & CRISPRi & 1 & --\\
LINCS L1000 \citep{subramanian2017} & 48 line and reagent sets & shRNA, knockout & 48 & --\\
\addlinespace[2pt]
\multicolumn{5}{@{}l}{\textit{Held out after the freeze to choose the blind test entry}}\\
mSWI/SNF screen \citep{otto2023} & MOLM-13 & Knockout & 1 & 0.03\\
Chromatin screen \citep{aryal2026} & MOLM-13 & Knockout & 1 & 0.01\\
Perturb-CITE-seq$^\dagger$ \citep{frangieh2021} & Patient-derived melanoma & Knockout & 1 & 0.22\\
Neuron CRISPRi$^\dagger$ \citep{tian2021} & iPSC-derived neurons & CRISPRi & 1 & 0.03\\
\bottomrule
\end{tabularx}
\end{table}

\FloatBarrier
\clearpage
\section{All Official Validation Submissions}
\label{app:subs}

\tab{tab:subs} lists all 47 official validation submissions in the order they were made, with the score, the rank that the score takes among the final entries of the validation leaderboard, and the six scaled metrics. The first six came from the warm-up and the rest from \px, and the shading shows at a glance which metric each step moved.

{\footnotesize
\setlength{\tabcolsep}{2pt}
\setlength{\LTleft}{0pt}\setlength{\LTright}{0pt}
\newcolumntype{L}[1]{>{\raggedright\arraybackslash}p{\dimexpr#1\linewidth-2\tabcolsep\relax}}
\newcolumntype{Q}[1]{>{\raggedleft\arraybackslash}p{\dimexpr#1\linewidth-2\tabcolsep\relax}}
\begin{longtable}{Q{.03}L{.06}L{.284}Q{.09}Q{.07}Q{.085}Q{.0635}Q{.0635}Q{.0635}Q{.0635}Q{.0635}Q{.0635}}
\caption{\textbf{All 47 validation submissions.} The rule separates warm-up entries from \px entries. Entries 5 and 7 repeat entry 3. $\star$: new best. Rank: where the score places on the leaderboard of 1{,}458 teams. Rank \%: that place as a share of the board. Arrows give the direction of improvement. Each column is shaded by rank, from its worst value in white to its best in blue, and its best value is in bold.}\label{tab:subs}\\
\toprule
\# & Day & Entry & Score$\uparrow$ & Rank$\downarrow$ & Rank \%$\downarrow$ & pds$\uparrow$ & mse$\uparrow$ & nmae$\uparrow$ & fid$\uparrow$ & reach$\uparrow$ & jac$\uparrow$\\
\midrule\endfirsthead
\toprule
\# & Day & Entry & Score$\uparrow$ & Rank$\downarrow$ & Rank \%$\downarrow$ & pds$\uparrow$ & mse$\uparrow$ & nmae$\uparrow$ & fid$\uparrow$ & reach$\uparrow$ & jac$\uparrow$\\
\midrule\endhead
1 & 0 & \texttt{lfc\_transfer\_sim v1} & \cellcolor[HTML]{F8FBFE}$\star$0.1041 & \cellcolor[HTML]{F8FBFE}608 & \cellcolor[HTML]{F8FBFE}41.7 & \cellcolor[HTML]{FAFCFE}0.452 & \cellcolor[HTML]{F6F9FE}0.000 & \cellcolor[HTML]{FAFCFE}0.074 & \cellcolor[HTML]{ECF3FC}$-$0.025 & \cellcolor[HTML]{FAFCFE}0.132 & \cellcolor[HTML]{F4F8FE}$-$0.008\\
2 & 1 & \texttt{delta\_v2} & \cellcolor[HTML]{FAFCFE}0.1037 & \cellcolor[HTML]{FAFCFE}609 & \cellcolor[HTML]{FAFCFE}41.8 & \cellcolor[HTML]{FDFEFF}0.447 & \cellcolor[HTML]{F6F9FE}0.000 & \cellcolor[HTML]{A2C5F2}0.118 & \cellcolor[HTML]{FDFEFF}$-$0.094 & \cellcolor[HTML]{EAF2FC}0.158 & \cellcolor[HTML]{F4F8FE}$-$0.008\\
3 & 1 & \texttt{xfer\_v2} & \cellcolor[HTML]{F3F8FD}$\star$0.1081 & \cellcolor[HTML]{F3F8FD}596 & \cellcolor[HTML]{F3F8FD}40.9 & \cellcolor[HTML]{F3F8FD}0.474 & \cellcolor[HTML]{F6F9FE}0.000 & \cellcolor[HTML]{9CC2F1}0.119 & \cellcolor[HTML]{F8FBFE}$-$0.091 & \cellcolor[HTML]{EFF5FD}0.156 & \cellcolor[HTML]{FAFCFE}$-$0.009\\
4 & 2 & \texttt{xfer\_eb\_f1} & \cellcolor[HTML]{FFFFFF}0.0901 & \cellcolor[HTML]{FFFFFF}652 & \cellcolor[HTML]{FFFFFF}44.7 & \cellcolor[HTML]{F8FBFE}0.465 & \cellcolor[HTML]{F6F9FE}0.000 & \cellcolor[HTML]{93BCF0}\textbf{0.128} & \cellcolor[HTML]{FFFFFF}$-$0.193 & \cellcolor[HTML]{E8F0FC}0.159 & \cellcolor[HTML]{FFFFFF}$-$0.018\\
5 & 2 & \texttt{xfer\_v2} & \cellcolor[HTML]{F3F8FD}0.1081 & \cellcolor[HTML]{F3F8FD}596 & \cellcolor[HTML]{F3F8FD}40.9 & \cellcolor[HTML]{F3F8FD}0.474 & \cellcolor[HTML]{F6F9FE}0.000 & \cellcolor[HTML]{9CC2F1}0.119 & \cellcolor[HTML]{F8FBFE}$-$0.091 & \cellcolor[HTML]{EFF5FD}0.156 & \cellcolor[HTML]{FAFCFE}$-$0.009\\
6 & 3 & \texttt{d4\_nce01\_eb01} & \cellcolor[HTML]{FDFEFF}0.0917 & \cellcolor[HTML]{FDFEFF}649 & \cellcolor[HTML]{FDFEFF}44.5 & \cellcolor[HTML]{FFFFFF}0.375 & \cellcolor[HTML]{F6F9FE}0.000 & \cellcolor[HTML]{F1F6FD}0.094 & \cellcolor[HTML]{F3F8FD}$-$0.051 & \cellcolor[HTML]{F8FBFE}0.139 & \cellcolor[HTML]{F0F6FD}$-$0.006\\
\midrule
7 & 5 & \texttt{xfer\_v2} & \cellcolor[HTML]{F3F8FD}0.1081 & \cellcolor[HTML]{F3F8FD}596 & \cellcolor[HTML]{F3F8FD}40.9 & \cellcolor[HTML]{F3F8FD}0.474 & \cellcolor[HTML]{F6F9FE}0.000 & \cellcolor[HTML]{9CC2F1}0.119 & \cellcolor[HTML]{F8FBFE}$-$0.091 & \cellcolor[HTML]{EFF5FD}0.156 & \cellcolor[HTML]{FAFCFE}$-$0.009\\
8 & 5 & \texttt{gen4\_peer5\_v3selnorm\_v1} & \cellcolor[HTML]{ECF3FC}$\star$0.1323 & \cellcolor[HTML]{ECF3FC}517 & \cellcolor[HTML]{ECF3FC}35.5 & \cellcolor[HTML]{EFF5FD}0.528 & \cellcolor[HTML]{F6F9FE}0.000 & \cellcolor[HTML]{98BFF1}0.121 & \cellcolor[HTML]{E0ECFB}$-$0.017 & \cellcolor[HTML]{E3EEFB}0.167 & \cellcolor[HTML]{F0F6FD}$-$0.006\\
9 & 7 & \texttt{gen0\_peer7\_wlo10\_c11...} & \cellcolor[HTML]{EAF2FC}$\star$0.1339 & \cellcolor[HTML]{EAF2FC}508 & \cellcolor[HTML]{EAF2FC}34.8 & \cellcolor[HTML]{ECF3FC}0.611 & \cellcolor[HTML]{F6F9FE}0.000 & \cellcolor[HTML]{95BDF0}0.123 & \cellcolor[HTML]{EFF5FD}$-$0.031 & \cellcolor[HTML]{FDFEFF}0.096 & \cellcolor[HTML]{95BDF0}0.005\\
10 & 8 & \texttt{es\_lsa\_a10\_det} & \cellcolor[HTML]{EFF5FD}0.1243 & \cellcolor[HTML]{EFF5FD}545 & \cellcolor[HTML]{EFF5FD}37.4 & \cellcolor[HTML]{EAF2FC}0.680 & \cellcolor[HTML]{EAF2FC}0.043 & \cellcolor[HTML]{FFFFFF}0.024 & \cellcolor[HTML]{F1F6FD}$-$0.048 & \cellcolor[HTML]{FFFFFF}0.038 & \cellcolor[HTML]{93BCF0}\textbf{0.008}\\
11 & 9 & \texttt{gen0\_peer1\_pw\_neara26} & \cellcolor[HTML]{E5EFFB}$\star$0.1787 & \cellcolor[HTML]{E5EFFB}291 & \cellcolor[HTML]{E5EFFB}20.0 & \cellcolor[HTML]{D6E6F9}0.727 & \cellcolor[HTML]{E8F0FC}0.113 & \cellcolor[HTML]{B3D0F4}0.109 & \cellcolor[HTML]{E9F1FC}$-$0.020 & \cellcolor[HTML]{F6F9FE}0.142 & \cellcolor[HTML]{9BC1F1}0.001\\
12 & 13 & \texttt{champ\_TxA70} & \cellcolor[HTML]{E8F0FC}0.1769 & \cellcolor[HTML]{E8F0FC}299 & \cellcolor[HTML]{E8F0FC}20.5 & \cellcolor[HTML]{E8F0FC}0.688 & \cellcolor[HTML]{E5EFFB}0.130 & \cellcolor[HTML]{A6C8F3}0.114 & \cellcolor[HTML]{E0ECFB}$-$0.017 & \cellcolor[HTML]{F3F8FD}0.146 & \cellcolor[HTML]{9BC1F1}0.001\\
13 & 16 & \texttt{champ\_spec130} & \cellcolor[HTML]{E3EEFB}$\star$0.1902 & \cellcolor[HTML]{E3EEFB}252 & \cellcolor[HTML]{E3EEFB}17.3 & \cellcolor[HTML]{D0E2F8}0.731 & \cellcolor[HTML]{E3EEFB}0.157 & \cellcolor[HTML]{AECDF4}0.110 & \cellcolor[HTML]{E5EFFB}$-$0.018 & \cellcolor[HTML]{E5EFFB}0.160 & \cellcolor[HTML]{9BC1F1}0.001\\
14 & 16 & \texttt{champ\_specauto} & \cellcolor[HTML]{E0ECFB}$\star$0.1934 & \cellcolor[HTML]{E0ECFB}235 & \cellcolor[HTML]{E0ECFB}16.1 & \cellcolor[HTML]{CDE0F8}0.732 & \cellcolor[HTML]{DFEBFB}0.172 & \cellcolor[HTML]{D0E2F8}0.103 & \cellcolor[HTML]{E0ECFB}$-$0.017 & \cellcolor[HTML]{E0ECFB}0.169 & \cellcolor[HTML]{9BC1F1}0.001\\
15 & 17 & \texttt{champ\_specauto\_gA} & \cellcolor[HTML]{DEEBFA}$\star$0.1964 & \cellcolor[HTML]{DEEBFA}223 & \cellcolor[HTML]{DEEBFA}15.3 & \cellcolor[HTML]{CDE0F8}0.732 & \cellcolor[HTML]{DFEBFB}0.172 & \cellcolor[HTML]{E5EFFB}0.101 & \cellcolor[HTML]{DCE9FA}$-$0.016 & \cellcolor[HTML]{DDEAFA}0.189 & \cellcolor[HTML]{C7DCF7}0.000\\
16 & 17 & \texttt{champ\_specauto\_cfg2\_gA} & \cellcolor[HTML]{D9E8FA}$\star$0.1978 & \cellcolor[HTML]{DBE8FA}220 & \cellcolor[HTML]{DBE8FA}15.1 & \cellcolor[HTML]{DFEBFB}0.719 & \cellcolor[HTML]{DBE8FA}0.184 & \cellcolor[HTML]{BAD4F5}0.108 & \cellcolor[HTML]{D9E8FA}$-$0.013 & \cellcolor[HTML]{DDEAFA}0.189 & \cellcolor[HTML]{C7DCF7}0.000\\
17 & 18 & \texttt{r22\_spread\_all10} & \cellcolor[HTML]{DCE9FA}0.1973 & \cellcolor[HTML]{DBE8FA}220 & \cellcolor[HTML]{DBE8FA}15.1 & \cellcolor[HTML]{DFEBFB}0.719 & \cellcolor[HTML]{DBE8FA}0.184 & \cellcolor[HTML]{E5EFFB}0.101 & \cellcolor[HTML]{E9F1FC}$-$0.020 & \cellcolor[HTML]{D9E8FA}0.200 & \cellcolor[HTML]{C7DCF7}0.000\\
18 & 18 & \texttt{r22\_lzkey\_k8} & \cellcolor[HTML]{CBDFF8}$\star$0.2046 & \cellcolor[HTML]{CBDFF8}192 & \cellcolor[HTML]{CEE0F8}13.2 & \cellcolor[HTML]{D6E6F9}0.727 & \cellcolor[HTML]{D5E5F9}0.189 & \cellcolor[HTML]{C8DDF7}0.104 & \cellcolor[HTML]{C4DBF7}$-$0.006 & \cellcolor[HTML]{D1E3F9}0.213 & \cellcolor[HTML]{C7DCF7}0.000\\
19 & 19 & \texttt{r23\_wfm06cap\_k16} & \cellcolor[HTML]{D1E3F9}0.2041 & \cellcolor[HTML]{D2E3F9}194 & \cellcolor[HTML]{D2E3F9}13.3 & \cellcolor[HTML]{D6E6F9}0.727 & \cellcolor[HTML]{CEE0F8}0.194 & \cellcolor[HTML]{F3F8FD}0.093 & \cellcolor[HTML]{C4DBF7}$-$0.006 & \cellcolor[HTML]{CBDFF8}0.215 & \cellcolor[HTML]{C7DCF7}0.000\\
20 & 19 & \texttt{r22\_lzkey\_k8\_hi125} & \cellcolor[HTML]{D7E6F9}0.1990 & \cellcolor[HTML]{D7E6F9}217 & \cellcolor[HTML]{D7E6F9}14.9 & \cellcolor[HTML]{D6E6F9}0.727 & \cellcolor[HTML]{D0E2F8}0.192 & \cellcolor[HTML]{FDFEFF}0.069 & \cellcolor[HTML]{C4DBF7}$-$0.006 & \cellcolor[HTML]{D7E6F9}0.211 & \cellcolor[HTML]{C7DCF7}0.000\\
21 & 20 & \texttt{r22\_lzk8\_fmens5} & \cellcolor[HTML]{D5E5F9}0.2030 & \cellcolor[HTML]{D5E5F9}199 & \cellcolor[HTML]{D5E5F9}13.6 & \cellcolor[HTML]{E3EEFB}0.718 & \cellcolor[HTML]{D7E6F9}0.186 & \cellcolor[HTML]{BAD4F5}0.108 & \cellcolor[HTML]{C4DBF7}$-$0.006 & \cellcolor[HTML]{D1E3F9}0.213 & \cellcolor[HTML]{C7DCF7}0.000\\
22 & 21 & \texttt{r25\_ptw\_cw} & \cellcolor[HTML]{CEE0F8}0.2044 & \cellcolor[HTML]{CFE1F8}193 & \cellcolor[HTML]{CEE0F8}13.2 & \cellcolor[HTML]{DCE9FA}0.725 & \cellcolor[HTML]{D2E3F9}0.190 & \cellcolor[HTML]{C2D9F7}0.106 & \cellcolor[HTML]{D5E5F9}$-$0.007 & \cellcolor[HTML]{D5E5F9}0.212 & \cellcolor[HTML]{C7DCF7}0.000\\
23 & 21 & \texttt{sub\_gt16\_s13} & \cellcolor[HTML]{D1E3F9}0.2041 & \cellcolor[HTML]{CFE1F8}193 & \cellcolor[HTML]{CEE0F8}13.2 & \cellcolor[HTML]{E5EFFB}0.709 & \cellcolor[HTML]{CBDFF8}0.199 & \cellcolor[HTML]{C4DBF7}0.105 & \cellcolor[HTML]{A5C7F2}$-$0.004 & \cellcolor[HTML]{CBDFF8}0.215 & \cellcolor[HTML]{C7DCF7}0.000\\
24 & 22 & \texttt{r26\_zkey\_fmmix} & \cellcolor[HTML]{C9DEF8}$\star$0.2122 & \cellcolor[HTML]{C9DEF8}167 & \cellcolor[HTML]{C9DEF8}11.5 & \cellcolor[HTML]{C2D9F7}0.748 & \cellcolor[HTML]{C9DEF8}0.207 & \cellcolor[HTML]{E5EFFB}0.101 & \cellcolor[HTML]{C4DBF7}$-$0.006 & \cellcolor[HTML]{AFCDF4}0.222 & \cellcolor[HTML]{C7DCF7}0.000\\
25 & 22 & \texttt{r28\_clusterproj} & \cellcolor[HTML]{C2D9F7}$\star$0.2170 & \cellcolor[HTML]{C2D9F7}160 & \cellcolor[HTML]{C4DBF7}11.0 & \cellcolor[HTML]{C0D8F6}0.750 & \cellcolor[HTML]{BDD6F6}0.238 & \cellcolor[HTML]{D0E2F8}0.103 & \cellcolor[HTML]{C4DBF7}$-$0.006 & \cellcolor[HTML]{C5DBF7}0.216 & \cellcolor[HTML]{C7DCF7}0.000\\
26 & 23 & \texttt{cp\_evtx} & \cellcolor[HTML]{C0D8F6}$\star$0.2177 & \cellcolor[HTML]{C0D8F6}158 & \cellcolor[HTML]{C0D8F6}10.8 & \cellcolor[HTML]{BDD6F6}0.756 & \cellcolor[HTML]{AFCDF4}0.247 & \cellcolor[HTML]{F6F9FE}0.091 & \cellcolor[HTML]{C4DBF7}$-$0.006 & \cellcolor[HTML]{C0D8F6}0.218 & \cellcolor[HTML]{C7DCF7}0.000\\
27 & 23 & \texttt{cp\_final\_allctx} & \cellcolor[HTML]{C4DBF7}0.2168 & \cellcolor[HTML]{C5DBF7}161 & \cellcolor[HTML]{C4DBF7}11.0 & \cellcolor[HTML]{C4DBF7}0.745 & \cellcolor[HTML]{C4DBF7}0.230 & \cellcolor[HTML]{A8C9F3}0.112 & \cellcolor[HTML]{D5E5F9}$-$0.007 & \cellcolor[HTML]{B4D0F5}0.220 & \cellcolor[HTML]{C7DCF7}0.000\\
28 & 24 & \texttt{nx\_pfizer\_all} & \cellcolor[HTML]{BCD5F6}$\star$0.2196 & \cellcolor[HTML]{BCD5F6}155 & \cellcolor[HTML]{BCD5F6}10.6 & \cellcolor[HTML]{C7DCF7}0.743 & \cellcolor[HTML]{B2CFF4}0.246 & \cellcolor[HTML]{AECDF4}0.110 & \cellcolor[HTML]{B2CFF4}$-$0.005 & \cellcolor[HTML]{A7C8F3}0.224 & \cellcolor[HTML]{C7DCF7}0.000\\
29 & 24 & \texttt{wl\_c3\_w80pcgen\_lock13} & \cellcolor[HTML]{C7DCF7}0.2162 & \cellcolor[HTML]{C5DBF7}161 & \cellcolor[HTML]{C4DBF7}11.0 & \cellcolor[HTML]{B5D1F5}0.773 & \cellcolor[HTML]{C2D9F7}0.234 & \cellcolor[HTML]{F8FBFE}0.080 & \cellcolor[HTML]{C4DBF7}$-$0.006 & \cellcolor[HTML]{C5DBF7}0.216 & \cellcolor[HTML]{C7DCF7}0.000\\
30 & 25 & \texttt{nr\_c2} & \cellcolor[HTML]{B4D0F5}$\star$0.2224 & \cellcolor[HTML]{B4D0F5}145 & \cellcolor[HTML]{B4D0F5}9.9 & \cellcolor[HTML]{B9D3F5}0.769 & \cellcolor[HTML]{B6D2F5}0.243 & \cellcolor[HTML]{B3D0F4}0.109 & \cellcolor[HTML]{C4DBF7}$-$0.006 & \cellcolor[HTML]{C0D8F6}0.218 & \cellcolor[HTML]{C7DCF7}0.000\\
31 & 25 & \texttt{nr\_sst30g} & \cellcolor[HTML]{BCD5F6}0.2196 & \cellcolor[HTML]{BCD5F6}155 & \cellcolor[HTML]{BCD5F6}10.6 & \cellcolor[HTML]{C9DEF8}0.742 & \cellcolor[HTML]{B4D0F5}0.245 & \cellcolor[HTML]{AACBF3}0.111 & \cellcolor[HTML]{B2CFF4}$-$0.005 & \cellcolor[HTML]{A0C4F2}0.225 & \cellcolor[HTML]{C7DCF7}0.000\\
32 & 26 & \texttt{tilt\_pc100} & \cellcolor[HTML]{B2CFF4}$\star$0.2236 & \cellcolor[HTML]{B0CEF4}142 & \cellcolor[HTML]{B0CEF4}9.7 & \cellcolor[HTML]{A7C8F3}0.783 & \cellcolor[HTML]{BBD5F6}0.239 & \cellcolor[HTML]{C0D8F6}0.107 & \cellcolor[HTML]{D5E5F9}$-$0.007 & \cellcolor[HTML]{B9D3F5}0.219 & \cellcolor[HTML]{C7DCF7}0.000\\
33 & 26 & \texttt{js\_dx\_cfg3} & \cellcolor[HTML]{B6D2F5}0.2216 & \cellcolor[HTML]{B6D2F5}147 & \cellcolor[HTML]{B6D2F5}10.1 & \cellcolor[HTML]{BBD5F6}0.759 & \cellcolor[HTML]{C0D8F6}0.237 & \cellcolor[HTML]{A2C5F2}0.118 & \cellcolor[HTML]{A5C7F2}$-$0.004 & \cellcolor[HTML]{B9D3F5}0.219 & \cellcolor[HTML]{C7DCF7}0.000\\
34 & 27 & \texttt{progbio\_qg4w} & \cellcolor[HTML]{B9D3F5}0.2204 & \cellcolor[HTML]{B9D3F5}153 & \cellcolor[HTML]{B9D3F5}10.5 & \cellcolor[HTML]{B2CFF4}0.776 & \cellcolor[HTML]{C7DCF7}0.226 & \cellcolor[HTML]{BAD4F5}0.108 & \cellcolor[HTML]{C4DBF7}$-$0.006 & \cellcolor[HTML]{C0D8F6}0.218 & \cellcolor[HTML]{C7DCF7}0.000\\
35 & 27 & \texttt{pg2\_pc100} & \cellcolor[HTML]{AFCDF4}$\star$0.2240 & \cellcolor[HTML]{B0CEF4}142 & \cellcolor[HTML]{B0CEF4}9.7 & \cellcolor[HTML]{A7C8F3}0.783 & \cellcolor[HTML]{B9D3F5}0.241 & \cellcolor[HTML]{BAD4F5}0.108 & \cellcolor[HTML]{C4DBF7}$-$0.006 & \cellcolor[HTML]{B9D3F5}0.219 & \cellcolor[HTML]{C7DCF7}0.000\\
36 & 28 & \texttt{t1} & \cellcolor[HTML]{AACBF3}$\star$0.2301 & \cellcolor[HTML]{AACBF3}132 & \cellcolor[HTML]{AACBF3}9.1 & \cellcolor[HTML]{93BCF0}\textbf{0.792} & \cellcolor[HTML]{AACBF3}0.281 & \cellcolor[HTML]{EFF5FD}0.098 & \cellcolor[HTML]{B2CFF4}$-$0.005 & \cellcolor[HTML]{CBDFF8}0.215 & \cellcolor[HTML]{C7DCF7}0.000\\
37 & 28 & \texttt{t2} & \cellcolor[HTML]{A6C8F3}$\star$0.2322 & \cellcolor[HTML]{A3C6F2}125 & \cellcolor[HTML]{A3C6F2}8.6 & \cellcolor[HTML]{98BFF1}0.786 & \cellcolor[HTML]{A6C8F3}0.288 & \cellcolor[HTML]{E5EFFB}0.101 & \cellcolor[HTML]{B2CFF4}$-$0.005 & \cellcolor[HTML]{ADCCF4}0.223 & \cellcolor[HTML]{C7DCF7}0.000\\
38 & 29 & \texttt{t3b} & \cellcolor[HTML]{A3C6F2}$\star$0.2325 & \cellcolor[HTML]{A3C6F2}125 & \cellcolor[HTML]{A3C6F2}8.6 & \cellcolor[HTML]{ADCCF4}0.781 & \cellcolor[HTML]{A1C5F2}0.291 & \cellcolor[HTML]{E5EFFB}0.101 & \cellcolor[HTML]{A5C7F2}$-$0.004 & \cellcolor[HTML]{A0C4F2}0.225 & \cellcolor[HTML]{C7DCF7}0.000\\
39 & 29 & \texttt{t3a} & \cellcolor[HTML]{A1C5F2}$\star$0.2332 & \cellcolor[HTML]{A3C6F2}125 & \cellcolor[HTML]{A3C6F2}8.6 & \cellcolor[HTML]{95BDF0}0.787 & \cellcolor[HTML]{A1C5F2}0.291 & \cellcolor[HTML]{DBE8FA}0.102 & \cellcolor[HTML]{A5C7F2}$-$0.004 & \cellcolor[HTML]{A7C8F3}0.224 & \cellcolor[HTML]{C7DCF7}0.000\\
40 & 30 & \texttt{t6} & \cellcolor[HTML]{ADCCF4}0.2278 & \cellcolor[HTML]{ADCCF4}135 & \cellcolor[HTML]{ADCCF4}9.3 & \cellcolor[HTML]{B5D1F5}0.773 & \cellcolor[HTML]{ADCCF4}0.278 & \cellcolor[HTML]{ECF3FC}0.100 & \cellcolor[HTML]{B2CFF4}$-$0.005 & \cellcolor[HTML]{B2CFF4}0.221 & \cellcolor[HTML]{C7DCF7}0.000\\
41 & 30 & \texttt{t7} & \cellcolor[HTML]{9CC2F1}$\star$0.2335 & \cellcolor[HTML]{9CC2F1}124 & \cellcolor[HTML]{9CC2F1}8.5 & \cellcolor[HTML]{AACBF3}0.782 & \cellcolor[HTML]{9CC2F1}0.294 & \cellcolor[HTML]{D0E2F8}0.103 & \cellcolor[HTML]{A5C7F2}$-$0.004 & \cellcolor[HTML]{9AC0F1}0.226 & \cellcolor[HTML]{C7DCF7}0.000\\
42 & 31 & \texttt{t11} & \cellcolor[HTML]{A8C9F3}0.2312 & \cellcolor[HTML]{A8C9F3}128 & \cellcolor[HTML]{A8C9F3}8.8 & \cellcolor[HTML]{AFCDF4}0.777 & \cellcolor[HTML]{A8C9F3}0.287 & \cellcolor[HTML]{C8DDF7}0.104 & \cellcolor[HTML]{A5C7F2}$-$0.004 & \cellcolor[HTML]{A7C8F3}0.224 & \cellcolor[HTML]{C7DCF7}0.000\\
43 & 31 & \texttt{t9} & \cellcolor[HTML]{93BCF0}\textbf{$\star$0.2344} & \cellcolor[HTML]{94BDF0}\textbf{122} & \cellcolor[HTML]{95BDF0}\textbf{8.4} & \cellcolor[HTML]{A1C5F2}0.784 & \cellcolor[HTML]{97BEF0}\textbf{0.296} & \cellcolor[HTML]{DBE8FA}0.102 & \cellcolor[HTML]{98BFF1}\textbf{$-$0.003} & \cellcolor[HTML]{93BCF0}\textbf{0.228} & \cellcolor[HTML]{C7DCF7}0.000\\
44 & 32 & \texttt{t9g3} & \cellcolor[HTML]{9FC3F2}0.2334 & \cellcolor[HTML]{9CC2F1}124 & \cellcolor[HTML]{9CC2F1}8.5 & \cellcolor[HTML]{A1C5F2}0.784 & \cellcolor[HTML]{A1C5F2}0.291 & \cellcolor[HTML]{DBE8FA}0.102 & \cellcolor[HTML]{98BFF1}\textbf{$-$0.003} & \cellcolor[HTML]{95BDF0}0.227 & \cellcolor[HTML]{C7DCF7}0.000\\
45 & 32 & \texttt{t9m} & \cellcolor[HTML]{9AC0F1}0.2340 & \cellcolor[HTML]{9CC2F1}124 & \cellcolor[HTML]{9CC2F1}8.5 & \cellcolor[HTML]{9BC1F1}0.785 & \cellcolor[HTML]{97BEF0}\textbf{0.296} & \cellcolor[HTML]{D0E2F8}0.103 & \cellcolor[HTML]{98BFF1}\textbf{$-$0.003} & \cellcolor[HTML]{A7C8F3}0.224 & \cellcolor[HTML]{C7DCF7}0.000\\
46 & 33 & \texttt{t9\_x8} & \cellcolor[HTML]{95BDF0}0.2343 & \cellcolor[HTML]{94BDF0}\textbf{122} & \cellcolor[HTML]{95BDF0}\textbf{8.4} & \cellcolor[HTML]{9BC1F1}0.785 & \cellcolor[HTML]{97BEF0}\textbf{0.296} & \cellcolor[HTML]{DBE8FA}0.102 & \cellcolor[HTML]{98BFF1}\textbf{$-$0.003} & \cellcolor[HTML]{9AC0F1}0.226 & \cellcolor[HTML]{C7DCF7}0.000\\
47 & 33 & \texttt{t9\_x16} & \cellcolor[HTML]{98BFF1}0.2342 & \cellcolor[HTML]{98BFF1}123 & \cellcolor[HTML]{95BDF0}\textbf{8.4} & \cellcolor[HTML]{A1C5F2}0.784 & \cellcolor[HTML]{97BEF0}\textbf{0.296} & \cellcolor[HTML]{D0E2F8}0.103 & \cellcolor[HTML]{98BFF1}\textbf{$-$0.003} & \cellcolor[HTML]{9AC0F1}0.226 & \cellcolor[HTML]{C7DCF7}0.000\\
\bottomrule
\end{longtable}
}

\end{document}